\documentclass[onefignum,onetabnum]{siamonline220329}

\usepackage{lipsum}
\usepackage{amsfonts}
\usepackage{graphicx}
\usepackage{epstopdf}
\usepackage{algorithmic}
\ifpdf
  \DeclareGraphicsExtensions{.eps,.pdf,.png,.jpg}
\else
  \DeclareGraphicsExtensions{.eps}
\fi

\usepackage{enumitem}
\setlist[enumerate]{leftmargin=.5in}
\setlist[itemize]{leftmargin=.5in}

\newcommand{\R}{\mathbb{R}}

\newsiamremark{remark}{Remark}
\newsiamremark{hypothesis}{Hypothesis}
\crefname{hypothesis}{Hypothesis}{Hypotheses}
\newsiamthm{claim}{Claim}

\headers{Applications of No-Collision Transportation Maps in Manifold Learning}{E. Negrini and L. Nurbekyan}

\title{Applications of No-Collision Transportation Maps in Manifold Learning\thanks{Submitted to editors on 04/21/2023, accepted in revised form on 11/02/2023.
\funding{E. N. was supported by Simons Postdoctoral program at IPAM and DMS 1925919 and AFOSR MURI FA9550-21-1-0084.
L. N. was partially supported by AFOSR MURI Grant FA9550-18-1-0502.}}}

\author{Elisa Negrini\thanks{Department of Mathematics, University of California Los Angeles, Los Angeles, CA 
  (\email{enegrini@math.ucla.edu}, \url{https://sites.google.com/view/elisa-negrini}).}
\and Levon Nurbekyan \thanks{Department of Mathematics, Emory University, Atlanta, GA  
  (\email{lnurbek@emory.edu})}}

\usepackage{amsopn}

\ifpdf
\hypersetup{
  pdftitle={Applications of No-Collision Transportation Maps in Manifold Learning},
  pdfauthor={E. Negrini and L. Nurbekyan}
}
\fi

\begin{document}

\maketitle

\begin{abstract}
In this work, we investigate applications of no-collision transportation maps introduced by Nurbekyan et al. in 2020 in manifold learning for image data. Recently, there has been a surge in applying transportation-based distances and features for data representing motion-like or deformation-like phenomena. Indeed, comparing intensities at fixed locations often does not reveal the data structure. No-collision maps and distances developed in [Nurbekyan et al., 2020] are sensitive to geometric features similar to optimal transportation (OT) maps but much cheaper to compute due to the absence of optimization. In this work, we prove that no-collision distances provide an isometry between translations (respectively dilations) of a single probability measure and the translation (respectively dilation) vectors equipped with a Euclidean distance. Furthermore, we prove that no-collision transportation maps, as well as OT and linearized OT maps, do not in general provide an isometry for rotations.  The numerical experiments confirm our theoretical findings and show that no-collision distances achieve similar or better performance on several manifold learning tasks compared to other OT and Euclidean-based methods at a fraction of a computational cost.
\end{abstract}



\section{Introduction}
High-dimensional datasets can be very difficult to visualize and may contain redundant information. In particular, while two- or three-dimensional data can be plotted to show its inherent structure, equivalent high-dimensional plots are much less intuitive. A fundamental insight in data science is that high-dimensional data frequently reveals underlying low-dimensional structures. In fact, a common assumption in machine learning, known as the \textit{manifold hypothesis}, is that observed data lie on a low-dimensional manifold embedded in a high-dimensional space \cite{lee2007nonlinear, jones2008manifold, chen2012nonlinear, fefferman2016testing}. Manifold learning is a machine learning approach to non-linear dimension reduction based on the manifold hypothesis. Many manifold learning algorithms have been proposed in recent years including Isomap \cite{tenenbaum2000global}, Diffusion Maps \cite{coifman2006diffusion}, Multidimensional Scaling \cite{borg2005modern}, and Laplacian Eigenmaps \cite{belkin2003laplacian}. These methods have numerous applications in machine learning, neural networks, pattern recognition, image processing, and computer vision~\cite{ma2011manifold}.

While extremely successful for some applications, most nonlinear dimension reduction methods embed data in Euclidean spaces implicitly assuming that Euclidean distances between data points are semantically meaningful. This assumption, however, may not be valid in general. Indeed, in many cases comparing intensities at fixed locations may not reveal the data structure. For instance, in image classification translated copies of a single image can have large Euclidean distance even though they are semantically identical and should be assigned the same label.

Recent works demonstrate that optimal transportation (OT) and related distances are effective in capturing data structure in various applications yielding somewhat more ``semantically meaningful" features and distances. In particular, data corresponding to some physical or material displacement, where data points can be modeled as probability distributions, seem to be more amenable to transportation type methods~\cite{kolouri17,moosmuller2023linear,hamm22wassmap,kileel21manifold,cloninger23linearized,mathews2019molecular}.


In this work, we pursue this previous line of research interpreting images as probability distributions on the pixel grid and performing manifold learning on image data using no-collision distances. The latter are transportation-type distances introduced in~\cite{nurbekyan20nocollision}. We show that no-collision distances achieve similar or better performance compared to other types of OT distances at a fraction of computational cost. More specifically, we study theoretically and numerically the performance of the metric Multidimensional Scaling (MDS)~\cite{borg2005modern} algorithm on several examples of manifolds in the space of probability measures when the pairwise distance matrix is computed using no-collision distances. Such analysis has been carried out using classical MDS in~\cite{hamm22wassmap,cloninger23linearized} with Wasserstein (OT) and linearized Wasserstein distances (LOT), respectively. MDS aims to embed data in a low-dimensional space so that pairwise Euclidean distances between data points are preserved. In this work we use the \textit{metric} MDS, but similar results can be obtained using the \textit{classical} MDS. The main difference between the two methods is that the classical MDS minimizes the \textit{strain} function instead of the \textit{stress} function and has an analytic solution via truncated SVD. More details about these methods can be found in \cite{borg2005modern}. In what follows, we use the term ``MDS" for referring to the ``metric MDS."

We demonstrate that no-collision maps faithfully capture spatial displacements similar to OT maps. At the same time, unlike OT distances, no optimization is required for computing no-collision maps. More precisely, we prove that no-collision transportation maps accurately capture translations and dilations of a given probability measure; that is, the no-collision transportation map between a measure and its translation coincides with the translation map, and the no-collision transportation map between a measure and its dilation coincides with the dilation map.

Consequently, no-collision distances provide an isometry between translations of a single probability measure and the translation vectors equipped with a Euclidean distance. Similarly, no-collision distances provide an isometry between dilations of a single probability measure and dilation vectors equipped with an anisotropic Euclidean distance. Since MDS provides an isometric embedding of data in a Euclidean space, these previous results guarantee that MDS with no-collision distances recovers translations and suitably scaled dilations of a single probability measure up to rigid transformations. These results are analogous to the ones proven for OT distances in~\cite{hamm22wassmap}. In contrast, LOT distances do not generally provide an isometry for dilations unless the reference measure is itself a dilation.

Furthermore, we prove that no-collision transportation maps do not in general provide an isometry for rotations. More precisely, we provide an explicit example of a curve in the space of probability measures generated by rotating a single measure that does not admit an isometric image lying in a planar circle when equipped with no-collision distances. Interestingly, we prove that OT and LOT do not generally provide isometries for rotations either. The possibility of these distances recovering rotations via (classical) MDS was empirically investigated in~\cite{hamm22wassmap,moosmuller2023linear} with some experiments suggesting otherwise.

As mentioned above, the computation of no-collision distances does not require optimization~\cite{nurbekyan20nocollision}, and one expects faster algorithms when utilizing them instead of optimization-based distances. In our numerical experiments, we compare the performance of no-collision maps with those of OT~\cite{hamm22wassmap} and LOT~\cite{moosmuller2023linear}. We observe that no-collision distances perform similarly or better in terms of accurately recovering OT distances and manifold structures but are faster to compute.

The rest of the paper is organized as follows. Sections~\ref{sec:prior} and \ref{sec:background} discuss prior work and background on optimal and no-collision transportation maps, respectively. 
Next, in Section~\ref{Theory}, we prove that no-collision maps recover translations and suitably scaled dilations of a single measure up to rigid transformations. Furthermore, in Section~\ref{RotThm}, we prove that no-collision, OT, and LOT maps do not generally provide isometries for rotations. We present our numerical experiments in Section~\ref{Exper}.

\section{Prior work}\label{sec:prior}

Manifold learning is a non-linear dimension reduction technique based on the assumption that observed data lie on a low-dimensional manifold embedded in a higher-dimensional space. More specifically, assume $(Y,d_Y)$ is a (possibly high dimensional) metric space where we observe  data $\{y_1,\cdots, y_m\} \subset Y$. The goal is to find a metric space $(\Theta, d_{\Theta})$ such that, for some $\{\theta_1,\cdots, \theta_m\} \subset {\Theta}$ we have $d_Y(y_i,y_j) \approx d_{\Theta}(\theta_i, \theta_j)$ for all $i,j$. A common choice is to look for $(\Theta, d_{\Theta}) = (\mathbb{R}^k, |\cdot|)$ with small $k$.

Common dimension reduction techniques include Multidimensional Scaling (MDS) \cite{borg2005modern},  Isomap \cite{tenenbaum2000global}, and Diffusion Maps \cite{coifman2006diffusion}. 
MDS aims to embed data in a low-dimensional space so that pairwise Euclidean distances between data points are preserved. Isomap builds on MDS but instead of Euclidean distances between data points it preserves pairwise graph-geodesic distances revealing the geometric structure of the data. 
Finally, Diffusion Maps embed data in a lower-dimensional space through the spectral decomposition of the diffusion operator on the data. With a few exceptions~\cite{kileel21manifold}, these methods assume data features in a Euclidean space with underlying Euclidean distance generating graph-geodesics and differential operators.

A more recent idea of using optimal transportation distances in machine learning and data science proved successful in many instances such as automatic translation, shape reconstruction, multi-label classification, brain decoding, manifold learning and more (see for instance \cite{digne2014feature, kolouri2015transport, kolouri17, alvarez2018gromov, hamm22wassmap, moosmuller2023linear}). Moreover, the optimal transportation framework provides a powerful theoretical framework due to the substantial work on optimal transportation related to PDEs and other fields \cite{villani2009optimal,santambrogio2015optimal, peyre19computational,villani2021topics}.

Recall that the square Wasserstein distance between probability measures $\mu$ and $\nu$ is given by
\begin{equation}\label{W2}
    W_2^2(\mu, \nu) := \min_{T:\,\nu = T\sharp \mu} \int_{\R^d} |x-T(x)|^2 d\mu(x)
\end{equation}
where $\nu = T\sharp \mu$ means that $T$ \textit{transports} $\mu$ into $\nu$; that is, $\nu(B)=\mu(T^{-1}(B))$ for all Borel subsets $B$. The solution of \eqref{W2} is called an optimal transportation map.

In \cite{hamm22wassmap}, the authors propose a variant of the Isomap~\cite{tenenbaum2000global}, called Wassmap, which uses Wasserstein distances instead of Euclidean distances when computing parwise distance matrix for data. The authors prove that for manifolds generated by translations or dilations of a fixed probability measure, Wassmap recovers the translation set or a scaled version of the dilation set up to a rigid transformation.
While very effective, Wassmap relies on the computation of Wasserstein distance which can be computationally expensive. Indeed, for each distance computation one has to solve an optimization problem \eqref{W2}. Hence, computing pairwise distances between $m$ distributions amounts to solving $\binom{m}{2}$ optimization problems. In~\cite{hamm22wassmap} the authors propose to trade off accuracy of the embedding with speed of computation by utilizing approximations of the true Wasserstein distance.

Numerous modifications and approximation methods for $W_2$ distances appeared recently including Linearized Optimal Transportation (LOT) \cite{want13LOT, khurana2023supervised, moosmuller2023linear}, Cumulative Distribution Transform (CDT) \cite{park2018cumulative}, Radon-CDT \cite{kolouri2015radon}, Sinkhorn distances \cite{cuturi2013sinkhorn, altschuler2019massively,bonafini2021domain, lin2022efficiency}, multiscale methods \cite{schmitzer2016sparse, gerber2017multiscale}, and no-collision transportation maps \cite{nurbekyan20nocollision}. The primary goal of these methods is to mitigate the computational cost imposed by computing OT distance while retaining some of the appealing geometric features of OT maps.




LOT embeds the space of probability distributions in $L^2(\mu;\mathbb{R}^d)$, where $\mu$ is the reference measure for building LOT maps. More specifically, we associate $\nu \in \mathcal{P}(\mathbb{R}^d)$ with the solution of~\eqref{W2} denoted by $T_{\mu}^{\nu}$ and define
\begin{equation*}
    W^2_{2,\mu}(\nu_1,\nu_2)=\int_{\mathbb{R}^d}|T_\mu^{\nu_1}(x)-T_\mu^{\nu_2}(x)|^2 d\nu(x).
\end{equation*}

In~\cite{cloninger23linearized}, the authors use $W_{2,\mu}$ instead of $W_2$ in the MDS to speed up computations which still require solving $m$ Wasserstein distance computations to compare $m$ distributions. Moreover, the result is dependent on the choice of the reference measure, $\nu$.

CDT is the LOT framework for 1D probability measures. In contrast to the higher-dimensional LOT, the Euclidean distance between two transformed (embedded) signals is the exact square Wasserstein distance. Radon-CDT is an extension of the CDT framework to 2D density functions based on sliced Wasserstein distance \cite{bonneel2015sliced}. The idea of sliced Wasserstein distance is to compute infinitely many linear projections of the high-dimensional distribution to one-dimensional distributions and then compute the average of the Wasserstein distance between these one-dimensional representations.

In this work, we use no-collision maps introduced in~\cite{nurbekyan20nocollision}. These maps preserve geometric properties of OT maps, such as no-collision, but are much faster to compute since they do not require optimization. More specifically, the method is based on a partition of probability measures into parts with equal masses via slicing in a sequence of directions chosen beforehand. While not optimal in general, the suboptimality of no-collision maps is often not severe as can be seen in Section~\ref{Exper} and~\cite[Section 4]{nurbekyan20nocollision}. Additionally, we prove in Section~\ref{Theory} that these maps are optimal if the two measures are a translation or dilation of one another.

Partitions with equal masses were also used in~\cite{ajtai1984optimal,trillos2015rate} for proving optimal rates of convergence of empirical measures in OT distances. One difference is that no-collision maps operate on the continuum level and match points in the supports of two absolutely continuous measures. Additionally, the existence of these maps is guaranteed beyond domains with Lipschitz boundary: see Theorem~\ref{thm:maps_existence} and \cite[Theorem 3]{nurbekyan20nocollision}.

\section{Preliminaries on optimal and no-collision transportation maps}\label{sec:background}

Here we present key definitions and results about optimal transportation and no-collision maps that we use.

\subsection{Optimal transportation}

We denote by $\mathcal{P}(\R^d)$ the set of Borel probability measures on $\R^d$. Furthermore, we denote by
\begin{equation*}
    \mathcal{P}_2(\mathbb{R}^d)=\left\{\mu \in \mathcal{P}(\mathbb{R}^d)~:~\int_{\R^d} |x|^2 d\mu(x) < \infty \right\}.
\end{equation*}
Finally, $\mathcal{P}_{ac}(\R^d)$ denotes the set of Borel probability measures over $\mathbb{R}^d$ that are absolutely continuous with respect to the $d$-dimensional Lebesgue measure, and
\begin{equation*}
    \mathcal{P}_{ac,2}(\mathbb{R}^d)=\left\{\mu \in \mathcal{P}_{ac}(\mathbb{R}^d)~:~\int_{\R^d} |x|^2 d\mu(x) < \infty \right\}.
\end{equation*}
\begin{definition}
Let $\mu \in \mathcal{P}(\mathbb{R}^d)$ and $T: \R^d \rightarrow \R^d$ is a Borel measurable map. The pushforward of $\mu$ via $T$, denoted by $T\sharp \mu$, is defined as
\begin{equation*}
T\sharp \mu(B) := \mu (T^{-1}(B)) \quad \forall B \subset \R^d \; \text{Borel}.    
\end{equation*}
For probability measures  $\mu,\,\nu\,\in \mathcal{P}(\R^d)$ we say that $T$ transports $\mu$ into $\nu$ if $T\sharp \mu = \nu$
\end{definition}


Assume that $c:\mathbb{R}^d\times \mathbb{R}^d \to \mathbb{R}$ is a Borel measurable nonnegative function so that $c(x,y)$ is the cost of transporting a unit mass from $x$ to $y$. Then the problem of optimally transporting $\mu$ to $\nu$ can be written as
\begin{equation}\label{OTM}
    \min_{T:\,\nu = T\sharp \mu} \int_{\R^d} c(x,T(x))d\mu(x).
\end{equation}
Minimizers of this problem are called \textit{optimal transportation maps}. In this paper, we consider the \textit{quadratic} cost $c(x,y)= |x-y|^2$ that yields
\begin{equation}\label{eq:W2}
    W^2_2(\mu, \nu) := \min_{T:\,\nu = T\sharp \mu} \int_{\R^d} |x-T(x)|^2 d\mu(x).
\end{equation}
$W_2(\mu,\nu)$ is called the \textit{square Wasserstein distance} between the measures $\mu,\nu$. Since among OT distances we only consider $W_2$ we refer to it simply as the Wasserstein distance.

In general,~\eqref{OTM} and~\eqref{eq:W2} might not have solutions or solutions might not be unique. The following theorem by Brenier provides conditions under which~\eqref{eq:W2} has a unique solution.
\begin{theorem}[Brenier \cite{brenier1991polar}]
Let $\mu, \nu \in \mathcal{P}_2(\R^d)$. If $\mu \in \mathcal{P}_{ac}(\mathbb{R}^d)$ then~\eqref{eq:W2} admits a $\mu$ a.e. unique solution. Moreover, there exists a convex function $\phi$ such that $T(x) = \nabla \phi(x)$ $\mu$ a.e..
\end{theorem}
Generalizations of this theorem to Riemannian manifolds and with more general cost functions can be found for example in \cite{mccann2001polar, ambrosio2013user, villani2009optimal}.

\subsection{No-collision transportation maps}\label{sec:no-collision}


Assume that $\mu\in \mathcal{P}_{ac}(\mathbb{R}^d)$ and consider
\begin{equation*}
    \mathcal{C}_0^\mu=\left\{ \Omega_0^\mu\right\},
\end{equation*}
where we denote by $\Omega_0^\mu=\mathbb{R}^d$. Next, put $b_1=0$, and let $s_1\in \mathbb{S}^{d-1}$. Since $\mu \in \mathcal{P}_{ac}(\mathbb{R}^d)$ there exists $h_1\in \mathbb{R}$ such that
\begin{equation*}
    \mu \left(x\cdot s_1 \leq h_1 \right)=\mu \left(x\cdot s_1 > h_1 \right)=\frac{1}{2}.
\end{equation*}
Then we set
$\Omega^\mu_{00}=\left\{x\cdot s_1 \leq h_1\right\},\quad \Omega^\mu_{01}=\left\{x\cdot s_1 > h_1\right\},\quad \mathcal{C}_1^\mu=\{\Omega^\mu_{00},\Omega^\mu_{01}\}.$\\
Next, we choose $b_2\in \{00,01\}$, and $s_2\in \mathbb{S}^{d-1}$. There exists $h_2 \in \mathbb{R}$ such that 
\begin{equation*}
    \mu \left(x\cdot s_2 \leq h_2 ~|~ \Omega^\mu_{b_2} \right)=\mu \left(x\cdot s_2 > h_2~ |~\Omega^\mu_{b_2}\right)=\frac{1}{2}.
\end{equation*}
Then we set
    $\Omega^\mu_{b_20}=\left\{x\cdot s_2 \leq h_2\right\} \cap \Omega^\mu_{b_2},\quad \Omega^\mu_{b_21}=\left\{x\cdot s_2 > h_2\right\} \cap \Omega^\mu_{b_2},$
and\\
$    \mathcal{C}_2^\mu=\mathcal{C}_1^\mu \setminus \Omega^\mu_{b_2} \cup \{\Omega^\mu_{b_20},\Omega^\mu_{b_21}\}.$\\
Going forward, at step $k+1$ we choose $b_{k+1}$ such that $\Omega^\mu_{b_{k+1}}\in \mathcal{C}_k^\mu$, and $s_{k+1} \in \mathbb{S}^{d-1}$. Then there exists $h_{k+1}\in \mathbb{R}$ such that
\begin{equation*}
    \mu \left(x\cdot s_{k+1} \leq h_{k+1} ~|~ \Omega^\mu_{b_{k+1}} \right)=\mu \left(x\cdot s_{k+1} > h_{k+1}~ |~\Omega^\mu_{b_{k+1}}\right)=\frac{1}{2}.
\end{equation*}
Then we set
\begin{equation*}
    \Omega^\mu_{b_{k+1}0}=\left\{x\cdot s_{k+1} \leq h_{k+1}\right\} \cap \Omega^\mu_{b_{k+1}},\quad \Omega^\mu_{b_{k+1}1}=\left\{x\cdot s_{k+1} > h_{k+1}\right\} \cap \Omega^\mu_{b_{k+1}},
\end{equation*}
and
\begin{equation*}
    \mathcal{C}_{k+1}^\mu=\mathcal{C}_k^\mu \setminus \Omega^\mu_{b_{k+1}} \cup \{\Omega^\mu_{b_{k+1}0},\Omega^\mu_{b_{k+1}1}\}.
\end{equation*}
We call the sequence $\mathcal{S}=\left((b_k,s_k) \right)_{k=1}^\infty$ a \textit{slicing schedule}.

To introduce no-collision distances and maps, we fix a slicing schedule $\mathcal{S}$ and we slice all $\mu \in \mathcal{P}_{ac}(\mathbb{R}^d)$ according $\mathcal{S}$. Loosely speaking, given $\mu,\nu$ and their partitions $\bigcup_{k=1}^\infty \mathcal{C}^\mu_{k}$ and $\bigcup_{k=1}^\infty \mathcal{C}^\mu_{\nu}$ we then seek to construct a map $t_{\mu}^\nu$ such that $t_\mu^\nu(\Omega^\mu_b) = \Omega^\nu_b$ (modulo sets of measure $0$) for all $\Omega^\mu_b \in \bigcup_{k=1}^\infty \mathcal{C}^\mu_{k}$ and $\Omega^\nu_b \in \bigcup_{k=1}^\infty \mathcal{C}^\nu_{k}$. Once such $t_{\mu}^\nu$ is found, we put
\begin{equation}\label{eq:W_Sp}
    W_{\mathcal{S},p}(\mu,\nu)=\left(\int_{\mathbb{R}^d}|t_{\mu}^\nu(x)-x|^p d\mu(x)\right)^{\frac{1}{p}}.
\end{equation}

The rigorous construction of $t_{\mu}^\nu$ is carried out in~\cite{nurbekyan20nocollision} as follows. Let $\mathcal{S}=\left((b_k,s_k) \right)_{k=1}^\infty$ be an arbitrary slicing schedule. For $\mu \in \mathcal{P}_{ac}(\mathbb{R}^d)$ denote by $\bigcup_{k=1}^\infty \mathcal{C}^\mu_k$ the partition of $\mu$ according to $\mathcal{S}$. Furthermore, we introduce $\ell^\mu_{k}:\mathbb{R}^d \to \mathbb{R}$ as follows
\begin{equation*}
    \ell^\mu_k(x)=\sum_{\alpha=0}^m \frac{i_\alpha}{3^\alpha},\quad x\in \Omega^\mu_{i_0i_1\cdots i_m},
\end{equation*}
for all $\Omega^\mu_{i_0i_1\cdots i_m} \in \mathcal{C}^\mu_k$, where $i_\alpha \in \{0,1\}$. According to~\cite[Lemma 1]{nurbekyan20nocollision}, $(\ell^\mu_k)_{k=0}^\infty$ is a bounded uniformly convergent sequence of measurable functions yielding a Borel measurable limit
\begin{equation*}
    \ell^\mu(x)=\lim\limits_{k \to \infty} \ell^\mu_k(x),\quad x\in \mathbb{R}^d.
\end{equation*}

\begin{definition}\label{def:calK}
Denote by $\mathcal{K}$ the set of all compactly supported probability measures $d\mu=f(x) dx \in \mathcal{P}_{ac}(\mathbb{R}^d)$ such that
 \begin{equation*}
 \mu\left(\partial(\mathrm{supp}(\mu)) \right)=0, \quad  c_\mu \leq f(x) \leq C_\mu, ~\mu ~\mbox{a.e.},
 \end{equation*}
for some constants $c_\mu,C_\mu>0$.
\end{definition}

The main theoretical results about the existence of no-collision transportation maps in~\cite{nurbekyan20nocollision} can be summarized in the following theorem.
\begin{theorem}[\cite{nurbekyan20nocollision}]\label{thm:maps_existence}
Suppose that $\mathcal{S}=\left((b_k,s_k)\right)_{k=1}^\infty$ is a slicing schedule such that
\begin{equation*}
\{s_k\}_{k=1}^\infty \subset \{e_i\}_{i=1}^d,     
\end{equation*}
where $\{e_i\}_{i=1}^d$ is an arbitrary basis in $\mathbb{R}^d$. Moreover, suppose that $\mathcal{S}$ is such that each set appearing during the partition gets partitioned in each of $\{e_i\}_{i=1}^d$ directions infinitely many times.

Then for every $\mu,\nu \in \mathcal{K}$ we have that
\begin{enumerate}
    \item there exists $E_\mu \subset \mathrm{int}(\mathrm{supp}(\mu))$ such that $\mu(E_\mu)=1$, and for every $x\in E_\mu$ there exits a unique $y \in \mathrm{int}(\mathrm{supp}(\nu))$ such that $\ell^\mu(x)=\ell^\nu(y)$,
    \item $t_{\mu}^\nu(x)=(\ell^\nu)^{-1}(\ell^\mu(x)),~x\in E_\mu$ is measurable, and $t_{\mu}^\nu\sharp \mu=\nu$,
    \item $(t_{\mu}^{\nu})^{-1}(\Omega^\nu_b)=E_\mu \cap \Omega^\mu_b$ for all $\Omega^\mu_b \in \bigcup_{k=1}^\infty \mathcal{C}^\mu_{k}$ and $\Omega^\nu_b \in \bigcup_{k=1}^\infty \mathcal{C}^\nu_{k}$,
    \item if $\Tilde{t}:\Tilde{E}\to \mathbb{R}^d$ is such that $\Tilde{E} \cap \Omega^\mu_b \subset \Tilde{t}^{-1}(\Omega^\nu_b)$ for all $\Omega^\mu_b \in \bigcup_{k=1}^\infty \mathcal{C}^\mu_{k}$ and $\Omega^\nu_b \in \bigcup_{k=1}^\infty \mathcal{C}^\nu_{k}$, and $\mu(\Tilde{E})=1$ then $t_{\mu}^\nu=\Tilde{t}$ $\mu$ a.e.
    \end{enumerate}
\end{theorem}

We use this theorem for computing no-collision transportation maps between translated and dilated measures. Furthermore, the metric properties of distances induced by no-collision maps can be summarized as follows.
\begin{proposition}\label{prp:W_Sp_metric}
    Let $\mathcal{S}$ be a slicing schedule such as in Theorem~\ref{thm:maps_existence}, $\sigma$ be the uniform probability measure on $[0,1]^d$, and $p\geq 1$. Then the following statements are true.
    \begin{enumerate}
        \item $\mu \mapsto t_\sigma^\mu$ is an injection from $\mathcal{K}$ into $L^p(\sigma;\mathbb{R}^d)$.
        \item $(\mathcal{K},W_{\mathcal{S},p})$ is a metric space.
    \end{enumerate}
\end{proposition}
\begin{proof}
    See Section~\ref{sm-sec:metric} in the Supplementary Material.
\end{proof}

\begin{remark}\label{slicing}
An instance of no-collision maps and distances depends on a slicing schedule we choose beforehand. In this paper, we consider only schedules such as in Theorem~\ref{thm:maps_existence}. In particular, all experiments are performed using successive horizontal and vertical cuts. The study of geometric properties and performance of no-collision maps in various manifold learning tasks depending on slicing schedules is an intriguing question that we plan to address elsewhere. Particularly interesting questions are related to quantitative estimates with respect to Wasserstein distances~\cite{delalande2023quantitative} and stability problems~\cite{moosmuller2023linear}.
\end{remark}

\subsection{Discretizing no-collision maps}

For discretizing no-collision maps we choose a finite slicing schedule $\mathcal{S}_N=\left( (b_k,s_k)\right)_{k=1}^N$ and partition given $\mu \in \mathcal{K}$ according to $\mathcal{S}_N$ obtaining sets $\mathcal{C}^\mu_N=\left\{ \Omega^\mu_b\right\}$. Then we build \textit{no-collision features} by taking either mass or geometric centers of $\Omega^\mu_b$; that is,
\begin{equation*}
    x^m_b(\mu)=\frac{1}{\mu(\Omega^\mu_b)} \int_{\Omega^\mu_b} x d\mu(x),\quad x^g_b(\mu)=\frac{1}{|\Omega^\mu_b|} \int_{\Omega^\mu_b} x dx,\quad \forall \Omega^\mu_b \in \mathcal{C}^\mu_N.
\end{equation*}
Hence, discrete no-collision maps map no-collision features; that is, for $\mu,\nu \in \mathcal{K}$ we define $\hat{t}^m:\left\{x^m_b(\mu)\right\} \to \left\{x^m_b(\nu)\right\}$ and $\hat{t}^g:\left\{x^g_b(\mu)\right\} \to \left\{x^g_b(\nu)\right\}$ as
\begin{equation*}
    \hat{t}^m(x^m_b(\mu))=x^m_b(\nu),\quad \hat{t}^g(x^g_b(\mu))=x^g_b(\nu),\quad \forall b.
\end{equation*}
Furthermore, no-collision distances can be discretized as
\begin{equation*}
    \hat{W}^m_{\mathcal{S}_N,p}(\mu,\nu)=\left(\sum_b \mu(\Omega^\mu_b)~|x^m_b(\mu)-x^m_b(\nu)|^p\right)^{\frac{1}{p}},
\end{equation*}
and
\begin{equation*}
    \hat{W}^g_{\mathcal{S}_N,p}(\mu,\nu)=\left(\sum_b \mu(\Omega^\mu_b)~|x^g_b(\mu)-x^g_b(\nu)|^p\right)^{\frac{1}{p}}.
\end{equation*}
In practice, choose $\mathcal{S}_N$ so that all $\Omega^\mu_b$ have the same mass and drop the factors $\mu(\Omega^\mu_b)$ thus computing $l_p$ distances between no-collision features. The algorithm for building no-collision features is as follows.

\begin{algorithm}[H]
\caption{Computing no-collision features}\label{alg:no-col}
\begin{algorithmic}
\REQUIRE{$N \geq 1$, $\mathcal{S}_N=\left( (b_k,s_k)\right)_{k=1}^N$, $\mu \in \mathcal{P}(\Omega)$.}
\STATE{ $k \gets 1$}
\STATE{ $\Omega^\mu_0 \gets \Omega$}
\STATE{ $\mathcal{C}^\mu \gets \left\{\Omega^\mu_0\right\}$}
\WHILE{$k <N$}
    \STATE{ \text{Find}~$h_k$~\text{s.t}~$\mu\left(x\cdot s_k\leq h_k~|~\Omega^\mu_{b_k} \right)=\mu\left(x\cdot s_k> h_k~|~\Omega^\mu_{b_k} \right)$}
    \STATE{ $\Omega^\mu_{b_k0}=\left\{ x\cdot s_k \leq h_k\right\} \cap \Omega^\mu_{b_k}$}
    \STATE{ $\Omega^\mu_{b_k1}=\left\{ x\cdot s_k > h_k\right\} \cap \Omega^\mu_{b_k}$}
    \STATE{ $\mathcal{C}^\mu \gets \mathcal{C}^\mu \setminus \Omega^\mu_{b_k} \cup \left\{\Omega^\mu_{b_k0},\Omega^\mu_{b_k1} \right\}$}
    \STATE{ $k \gets k+1$}
\ENDWHILE
\FOR{$\Omega^\mu_b \in \mathcal{C}^\mu$}
\STATE{ $x^m_b(\mu)=\frac{1}{\mu(\Omega^\mu_b)} \int_{\Omega^\mu_b} x d\mu(x)$}
\STATE{ $x^g_b(\mu)=\frac{1}{|\Omega^\mu_b|} \int_{\Omega^\mu_b} x dx$}
\ENDFOR
\end{algorithmic}
\end{algorithm}

\section{Translation and dilation manifolds}\label{Theory}

Here, we show that translation and dilation manifolds can be learned by no-collision maps just as with OT maps.

\subsection{Translation manifolds}\label{TransThm}

\begin{theorem}\label{thm:translation_isometry}
Assume that $\mu_0 \in \mathcal{K}$, and let $\mu_{\theta}=(x+\theta)\sharp \mu_0$ for $\theta \in \mathbb{R}^d$. Then for every slicing schedule $\mathcal{S}$ such as in Theorem~\ref{thm:maps_existence} we have that
\[
W_{\mathcal{S},p}(\mu_{\theta},\mu_{\theta'})=W_p(\mu_{\theta},\mu_{\theta'})=|\theta-\theta'|,\quad \forall \theta,\theta' \in \mathbb{R}^d,~p\geq 1.
\]
\end{theorem}
\begin{proof}
Firstly, note that it is enough to prove the identity for $\theta'=0$. Indeed, $\mu_{\theta}=(x+(\theta-\theta'))\sharp \mu_{\theta'}$. Next, let $\mathcal{S}=\left((b_k,s_k)\right)_{k=1}^\infty$ be a slicing schedule as in Theorem~\ref{thm:maps_existence}. Our goal is to compute $\ell^{\mu_\theta}$, and show that
\begin{equation*}
    \ell^{\mu_{\theta}}(x)=\ell^{\mu_0}(x-\theta),\quad \forall x\in \mathbb{R}^d.
\end{equation*}
We have that $\mathcal{C}^{\mu_0}_0=\{\Omega_0\}$, where $\Omega^{\mu_0}_0=\mathbb{R}^d$, and so
$\ell_0^{\mu_\theta}(x)=0,\quad \forall x\in \mathbb{R}^d.$\\
Next, let $h_1$ be such that
$\mu_0\left( \{x\cdot s_1\leq h_1\}\right)=\mu_0\left( \{x\cdot s_1> h_1\}\right)=\frac{1}{2}.$\\
Then we have that
$\Omega^{\mu_0}_{00}=\{x\cdot s_1\leq h_1\},\quad \Omega^{\mu_0}_{01}=\{x\cdot s_1> h_1\}.$\\
Note that
\begin{equation*}
    \begin{split}
    \mu_{\theta}\left( \{x\cdot s_1\leq h_1+\theta\cdot s_1\}\right)=&\mu_0 \left( \{(x+\theta)\cdot s_1\leq h_1+\theta\cdot s_1\}\right) \\
    =&\mu_0 \left( \{x\cdot s_1\leq h_1\}\right)=\frac{1}{2},
    \end{split}
\end{equation*}
and so
    $\mu_{\theta}\left( \{x\cdot s_1\leq h_1+\theta\cdot s_1\}\right)=\mu_{\theta}\left( \{x\cdot s_1 > h_1+\theta\cdot s_1\}\right)=\frac{1}{2}.$\\
This means that the first two slices for $\mu_\theta$ are
\[
\Omega^{\mu_\theta}_{00}=\{x\cdot s_1\leq h_1+\theta\cdot s_1\},\quad \Omega^{\mu_\theta}_{01}=\{x\cdot s_1> h_1+\theta \cdot s_1\}.
\]
But this means that
$\Omega^{\mu_\theta}_{00}=\Omega^{\mu_0}_{00}+\theta,\;\text{and}\; \Omega^{\mu_\theta}_{01}=\Omega^{\mu_0}_{01}+\theta.$\\
The arguments above apply to all partitions, and we obtain
\[
\Omega^{\mu_\theta}_{b}=\Omega^{\mu_0}_{b}+\theta
\]
for all $\Omega^{\mu_\theta}_b \in \bigcup_{k=1}^\infty \mathcal{C}^{\mu_\theta}_{k}$ and $\Omega^{\mu_0}_b \in \bigcup_{k=1}^\infty \mathcal{C}^{\mu_0}_{k}$. Hence $x\in \Omega^{\mu_\theta}_b$ if and only if $x-\theta \in \Omega^{\mu_0}_b$.\\ 
Since $\ell^{\mu_{\theta}}_k(x)$ depends only on $b$ for $x\in \Omega^{\mu_\theta}_b$ we obtain that
    $\ell^{\mu_\theta}_k(x)=\ell^{\mu_0}_k(x-\theta),\; \forall x \in \Omega^{\mu_\theta}_b,$
and so
   $ \ell^{\mu_\theta}_k(x)=\ell^{\mu_0}_k(x-\theta),\quad \forall x \in \mathbb{R}^d.$\\
Passing to the limit in $k$ we find that
    $\ell^{\mu_\theta}(x)=\ell^{\mu_0}(x-\theta),\quad \forall x \in \mathbb{R}^d.$
Taking into account Theorem~\ref{thm:maps_existence}, we obtain that $t_{\mu_0}^{\mu_\theta}(x)=x+\theta$ for $\mu_0$ a.e. $x$. Note that $t_{\mu_0}^{\mu_\theta}(x)=\nabla \left( \frac{|x|^2}{2}+\theta \cdot x \right)$, and $x\mapsto \frac{|x|^2}{2}+\theta \cdot x $ is convex. Hence by Brenier's theorem $t_{\mu_0}^{\mu_\theta}$ is the optimal transportation map from $\mu_0$ to $\mu_\theta$, and
\begin{equation*}
    W_{\mathcal{S},p}(\mu_\theta,\mu_0)=W_{p}(\mu_\theta,\mu_0)=\left(\int_{\mathbb{R}^d}|t_{\mu_0}^{\mu_\theta}(x)-x|^p d\mu_0(x)\right)^{\frac{1}{p}}=|\theta|.
\end{equation*}
\end{proof}

\begin{corollary}
Assume that $\mu_0 \in \mathcal{K}$, $\Theta \subset \mathbb{\R}^d$, $p\geq 1$, and $\mathcal{S}$ is a slicing schedule such as in Theorem~\ref{thm:maps_existence}. Then $(\{\mu_\theta\},W_{\mathcal{S},p})$ is isometric to $(\Theta,|\cdot|)$.
\end{corollary}

\subsection{Dilation manifolds}\label{DilThm}


Recall that the Hadamard product of $x=(x_1,x_2,\cdots,x_d)$ and $y=(y_1,y_2,\cdots,y_d)$ is defined as
   $ x \odot y = (x_1 y_1,x_2 y_2,\cdots,x_d y_d).$

\begin{theorem}\label{thm:dilation_isometry}
Assume that $\mu_0\in \mathcal{K}$, and let $\mu_\theta=\left(\theta \odot x \right)\sharp \mu_0$ for $\theta \in \mathbb{R}^d_+$. Then for every slicing schedule $\mathcal{S}$ such as in Theorem~\ref{thm:maps_existence} containing only directions parallel to coordinate axes we have that
\[
W_{\mathcal{S},2}(\mu_\theta,\mu_{\theta'})=W_{2}(\mu_\theta,\mu_{\theta'})=|c\odot \theta - c\odot \theta'|,\quad \forall \theta,\theta' \in \mathbb{R}^d,
\]
where
\[
c^2_j=\int_{\mathbb{R}^d} x_j^2 d\mu_0(x),\quad 1\leq j \leq d.
\]
\end{theorem}

\begin{proof}
Let $\mathcal{S}=((b_k,s_k))_{k=1}^\infty$ be a slicing such as in Theorem~\ref{thm:maps_existence} with $\{s_k\}$ parallel to coordinate axes, and $\theta,\theta'\in \mathbb{R}^d_+$. Our goal is to show that
\[
t_{\mu_\theta}^{\mu_{\theta'}}(x)=\left(\frac{\theta'_1}{\theta_1}x_1,\frac{\theta'_2}{\theta_2}x_2,\cdots,\frac{\theta'_d}{\theta_d}x_d \right),\quad\mu_\theta~\text{a.e.}. 
\]
We have that
$ \ell^{\mu_\theta}_0(x)=   \ell^{\mu_{\theta'}}_0(x)=0,\quad \forall x\in \mathbb{R}^d.$\\
Without loss of generality, assume that $s_1=(1,0,\cdots,0)$. Let $h_1$ be such that
\[
\mu_\theta\left( \{x_1\leq h_1\}\right)=\mu_\theta\left( \{x_1> h_1\}\right)=\frac{1}{2}.
\]
But then we have that $\mu_{\theta'}=\left(\frac{\theta'_1}{\theta_1} x_1,\frac{\theta'_2}{\theta_2} x_2,\cdots.\frac{\theta'_d}{\theta_d} x_d \right)\sharp \mu_\theta$, and so
\begin{equation*}
    \begin{split}
        \mu_{\theta'}\left(\left\{x_1 \leq \frac{\theta_1'}{\theta_1} h_1\right\}\right)=&\mu_\theta \left(\left\{\frac{\theta_1'}{\theta_1} x_1 \leq \frac{\theta_1'}{\theta_1} h_1\right\}\right)\\
        =&\mu_\theta \left(\left\{x_1 \leq h_1\right\}\right)=\frac{1}{2}.
    \end{split}
\end{equation*}
Hence,
 $   \mu_{\theta'}\left(\left\{x_1 \leq \frac{\theta_1'}{\theta_1} h_1\right\}\right)=\mu_{\theta'}\left(\left\{x_1 > \frac{\theta_1'}{\theta_1} h_1\right\}\right)=\frac{1}{2}.$\\
This means that
$\Omega^{\mu_\theta}_{00}=\{x_1\leq h_1\},\quad \Omega^{\mu_\theta}_{01}=\{x_1> h_1\},$
and
$\Omega^{\mu_{\theta'}}_{00}=\left\{x_1\leq \frac{\theta'_1}{\theta_1} h_1\right\}$,\\$ \Omega^{{\mu_\theta'}}_{01}=\left\{x_1> \frac{\theta'_1}{\theta_1} h_1\right\}.$
So the dilation map
\begin{equation*}
    (x_1,x_2,\cdots,x_d) \mapsto \left(\frac{\theta'_1}{\theta_1}x_1,\frac{\theta'_2}{\theta_2}x_2,\cdots,\frac{\theta'_d}{\theta_d}x_d \right)
\end{equation*}
maps $\Omega^{\mu_{\theta}}_{00}$ to $\Omega^{\mu_{\theta'}}_{00}$ and $\Omega^{\mu_{\theta}}_{01}$ to $\Omega^{\mu_{\theta'}}_{01}$. Similarly, we obtain that this dilation map maps $\Omega^{\mu_\theta}_{b}$ to $\Omega^{\mu_{\theta'}}_{b}$ for all $\Omega^{\mu_\theta}_b \in \bigcup_{k=1}^\infty \mathcal{C}^{\mu_\theta}_{k}$ and $\Omega^{\mu_0}_b \in \bigcup_{k=1}^\infty \mathcal{C}^{\mu_0}_{k}$. Again, since $\ell^{\mu_\theta}_k(x)$ depends only on $b$ for $x\in \Omega^{\mu_\theta}_b$ we obtain that
\begin{equation*}
    \ell^{\mu_{\theta}}_k(x)=\ell^{\mu_{\theta'}}_k\left(\frac{\theta_1'}{\theta_1}x_1,\frac{\theta_2'}{\theta_2}x_2,\cdots,\frac{\theta_d'}{\theta_d}x_d \right),\quad \forall x\in \Omega^{\mu_\theta}_b,
\end{equation*}
and so
\begin{equation*}
\ell^{\mu_{\theta}}_k(x)=\ell^{\mu_{\theta'}}_k\left(\frac{\theta_1'}{\theta_1}x_1,\frac{\theta_2'}{\theta_2}x_2,\cdots,\frac{\theta_d'}{\theta_d}x_d \right),\quad \forall x\in \mathbb{R}^d.
\end{equation*}
Passing to the limit in $k$ we find that
\begin{equation*}
\ell^{\mu_{\theta}}(x)=\ell^{\mu_{\theta'}}\left(\frac{\theta_1'}{\theta_1}x_1,\frac{\theta_2'}{\theta_2}x_2,\cdots,\frac{\theta_d'}{\theta_d}x_d \right),\quad \forall x\in \mathbb{R}^d.
\end{equation*}
Taking into account Theorem~\ref{thm:maps_existence} we obtain that
\begin{equation*}
    t_{\mu_\theta}^{\mu_{\theta'}}(x)=\left(\frac{\theta_1'}{\theta_1}x_1,\frac{\theta_2'}{\theta_2}x_2,\cdots,\frac{\theta_d'}{\theta_d}x_d \right){=\nabla \sum_{j=1}^d \frac{\theta'_j}{2\theta_j} x_j^2},\quad \mu_\theta~\text{a.e.}.
\end{equation*}
Since $x\mapsto \sum_{j=1}^d \frac{\theta'_j}{2\theta_j} x_j^2$ is convex, Brenier's theorem implies that $t_{\mu_\theta}^{\mu_{\theta'}}$ is the optimal transportation map from $\mu_\theta$ to $\mu_{\theta'}$, and
\begin{equation*}
    \begin{split}
 W_{\mathcal{S},2}^2(\mu_\theta,\mu_{\theta'})=& W_{2}^2(\mu_\theta,\mu_{\theta'})=\int_{\mathbb{R}^d} \left|t_{\mu_\theta}^{\mu_{\theta'}}(x)-x \right|^2 d\mu_\theta(x)=\int_{\mathbb{R}^d} \sum_{j=1}^d \left( \frac{\theta'_j}{\theta_j}x_j-x_j \right)^2 d\mu_\theta(x)\\
 =&\int_{\mathbb{R}^d} \sum_{j=1}^d \left( \theta'_jx_j-\theta_j x_j \right)^2 d\mu_0(x)=\sum_{j=1}^d c_j^2 (\theta'_j-\theta_j)^2.
    \end{split}
\end{equation*}
\end{proof}

\begin{corollary}
    Assume that $\mu_0 \in \mathcal{K}$, $\Theta \subset \mathbb{\R}^d_+$, and $\mathcal{S}$ is an arbitrary schedule with only slicing directions parallel to coordinate axes. Then $(\{\mu_\theta\},W_{\mathcal{S},2})$ is isometric to $\left(c\odot \Theta,|\cdot|\right)$, where
    \[
    c_j^2=\int_{\mathbb{R}^d} x_j^2 d\mu_0(x),\quad 1\leq j \leq d.
    \]
\end{corollary}


\section{Rotation manifolds}\label{RotThm}


In~\cite{hamm22wassmap} the authors investigated whether some structural information can be learned from pairwise distances of the subset
\begin{equation*}
    \left\{(Rx)\sharp \mu_0~:~R\in SO(d)\right\} \subset \mathcal{P}_2(\mathbb{R}^d).
\end{equation*}
For example, can one find an isometry between this subset and some representation of $SO(d)$ in a Euclidean space? One particular question considered in~\cite{hamm22wassmap} is as follows. Denote by $R_t$ the counter-clockwise rotation by angle $t$ around the origin; that is,
\[
R_tx=(x_1\cos t -x_2 \sin t, x_1 \sin t+x_2 \cos t).
\]
Then the question is whether one generically has that $\{(R_tx)\sharp \mu_0~:~t\in [0,2\pi]\}$ with $W_2$ distance is isometric to $\left( \partial B_r, |\cdot|\right)$ for some $r>0$, where $\partial B_r=\left\{x\in \mathbb{R}^2~:~|x|=r\right\}$. In~\cite{hamm22wassmap}, the authors perform a numerical experiment that recovers what looks like a circle when $\mu_0$ is the uniform measure over an ellipse. Similar question is considered in~\cite{cloninger23linearized}, where the $W_2$ is replaced by a $W_{2,\nu}$ -- the linearized $W_2$ distance with a reference measure $\nu$. 

Here we show that generically neither metric makes the set 
\[
\{(R_t x)\sharp \mu_0~:~t\in [0,2\pi]\}
\]
isometric to a planar circle.  

We achieve this by analytically computing the $W_2$ and $W_{2,\nu}$ distances between rotations of uniform probability measures over ellipses in plane as considered in a numerical experiments in~\cite{hamm22wassmap} and between rotations of Gaussians as considered in numerical experiments in~\cite{cloninger23linearized}. In fact, these examples are somewhat identical in terms of pairwise distances since OT maps and distances between two uniform measures over ellipses match, up to scaling, with OT maps and distances of corresponding Gaussian measures~\cite{gelbrich90} and ~\cite[Remarks 2.31-32]{peyre19computational}. 

Finally, we show that no-collision maps do not resolve the isometry problem either.

\subsection{Wasserstein distances}

We start with preliminary lemmas.

\begin{lemma}\label{lma:rotation}
Assume that $\mu_0 \in \mathcal{P}_2(\mathbb{R}^2)$, and $\mu_t=(R_{t}x)\sharp \mu_0$. Then we have that
\begin{equation*}
W_2(\mu_s,\mu_t)=W_2(\mu_0,\mu_{t-s}),\quad 0\leq s<t \leq 2\pi.    
\end{equation*}
\end{lemma}

\begin{proof}
Recall that for $\mu,\nu \in \mathcal{P}_2(\mathbb{R}^d)$ we denote by $\Pi(\mu,\nu)$ the set of all transportation plans with marginals $\mu,\nu$. Additionally, we denote by $\Pi_0(\mu,\nu)$ the set of all optimal transportation plans from $\mu$ to $\nu$. Let $\pi \in \mathcal{P}_2(\R^2\times \R^2)$. Then we define $\hat{\pi}=(R_sx,R_sy)\sharp \pi$. It is straightforward to check that $\pi=(R_{-s}x,R_{-s}y)\sharp \hat{\pi}$, and $\hat{\pi} \in \Pi(\mu_s,\mu_t)$ if and only if $\pi \in \Pi(\mu_0,\mu_{t-s})$. Furthermore, we have that
\begin{equation*}
\begin{split}
    \int_{\R^2\times \R^2} |x-y|^2 d\hat{\pi}(x,y)=&\int_{\R^2\times \R^2} |R_sx-R_sy|^2 d\pi(x,y)\\
    =&\int_{\R^2\times \R^2} |x-y|^2 d\pi(x,y)
\end{split}
\end{equation*}
Choosing $\pi \in \Pi_0(\mu_0,\mu_{t-s})$ we obtain that $W_2(\mu_s,\mu_t) \leq W_2(\mu_0,\mu_{t-s})$. Next, choosing $\hat{\pi}\in \Pi_0(\mu_s,\mu_{t})$ we obtain that $W_2(\mu_s,\mu_t) \geq W_2(\mu_0,\mu_{t-s})$. Hence $W_2(\mu_s,\mu_t) = W_2(\mu_0,\mu_{t-s})$, and $\pi \in \Pi_0(\mu_0,\mu_{t-s})$ if and only if $\hat{\pi}\in \Pi_0(\mu_s,\mu_t)$.
\end{proof}

\begin{lemma}\label{lma:ac_curve}
Assume that $\mu_0 \in \mathcal{P}_2(\mathbb{R}^2)$, and $\mu_t=(R_{t}x)\sharp \mu_0$. Then $\{\mu_t~:~t\in [0,2\pi]\}$ is an absolutely continuous curve with constant metric derivative.    
\end{lemma}
\begin{proof}
Let $0\leq s<t\leq 2\pi$. We have that $(R_{t-s}x)\sharp \mu_0=\mu_{t-s}$, and so
\begin{equation*}
    \begin{split}
        W_2^2(\mu_0,\mu_{t-s})\leq & \int_{\R^2} |R_{t-s}x-x|^2 d\mu_0(x)=4\sin^2 \frac{t-s}{2} \int_{\R^2} |x|^2\mu_0(x)\\
        \leq & c |t-s|^2.
    \end{split}
\end{equation*}
Invoking Lemma~\ref{lma:rotation} we obtain
$W_2(\mu_s,\mu_t)\leq c|t-s|,\quad 0\leq s<t \leq 2\pi.$\\
So $\{\mu_t~:~t\in [0,2\pi]\}$ is Lipschitz and therefore absolutely continuous. Applying \cite[Theorem 1.1.2]{ags08} we obtain that there exists
\begin{equation*}
    |\mu'_t|=\lim_{h\to 0} \frac{W_2(\mu_t,\mu_{t+h})}{|h|}
\end{equation*}
for Lebesgue a.e. $t\in [0,2\pi]$. Invoking Lemma~\ref{lma:rotation} again, we have that
\[
W_2(\mu_t,\mu_{t+h})=W_2(\mu_{t'},\mu_{t'+h}),\quad \forall t,t' \in [0,2\pi],
\]
and so if the metric derivative exists at one point it exists everywhere, and it is constant:
\begin{equation*}
    |\mu'_t|=\lim_{h\to 0} \frac{W_2(\mu_0,\mu_{h})}{|h|}=c,\quad \forall t\in [0,2\pi],
\end{equation*}
as desired
\end{proof}

\begin{proposition}\label{prp:analytic_distances}
Assume that $a,b>0$, and $\mu_0$ is the uniform probability measure over the elliptical domain
\begin{equation*}
    \mathcal{E}=\left\{(x_1,x_2)\in \R^2~:~\frac{(x_1-u_1)^2}{a^2}+\frac{(x_2-u_2)^2}{b^2} \leq 1\right\}.
\end{equation*}
Furthermore, assume that $\mu_t=(R_{t}x)\sharp \mu_0$. Then we have that
\begin{equation}\label{WssTrue}
\begin{split}
    &W_2^2(\mu_s,\mu_t)\\
    =&4|u|^2\sin^2 \frac{t-s}{2}+\frac{1}{2}\left(a^2+b^2-\sqrt{ (a^2+b^2)^2\cos^2 (t-s)+4a^2 b^2 \sin^2(t-s)} \right),
\end{split}
\end{equation}
for all $s,t \in  [0,2\pi]$, where $u=(u_1,u_2)$.
\end{proposition}
\begin{proof}
Taking into account Lemma~\ref{lma:rotation} we only need to prove \eqref{WssTrue} for $s=0$. From~\cite{gelbrich90} (see also~\cite[Remarks 2.31-32]{peyre19computational}) we have that
\begin{equation*}
    W_2^2(\mu_0,\mu_t)=|m_0-m_t|^2+\operatorname{tr}\left(\Sigma_0+\Sigma_t-2\left(\Sigma_0^{1/2}\Sigma_t \Sigma_0^{1/2}\right)^{1/2}\right),
\end{equation*}
where $m_t$ and $\Sigma_t$ are the mean and covariance of $\mu_t$, respectively. A direct computation yields
\begin{equation*}
    m_0=u,\quad \Sigma_0=\begin{pmatrix}
    \frac{a^2}{4}& 0\\
    0& \frac{b^2}{4}
    \end{pmatrix}.
\end{equation*}
Furthermore, we have that
   $ m_t=R_t m_0=R_t u,$\\
and
\begin{equation*}
    \Sigma_t=R_t \Sigma_0 R_{-t}=\begin{pmatrix}
    \frac{a^2 \cos^2 t+b^2 \sin^2 t}{4}& \frac{(a^2-b^2)\cos t \sin t}{4}\\
    \frac{(a^2-b^2)\cos t \sin t}{4} & \frac{a^2 \sin^2 t+b^2 \cos^2 t}{4}
    \end{pmatrix}.
\end{equation*}
To conclude the proof, we need to calculate $\operatorname{tr}\left(\Sigma_0^{1/2} \Sigma_t \Sigma_0^{1/2}  \right)^{1/2}$.\\Denoting by $M=\Sigma_0^{1/2} \Sigma_t \Sigma_0^{1/2}$ we have that
    $\operatorname{tr}(M^{1/2})=\left(\operatorname{tr}(M)+2 \sqrt{\det(M)} \right)^{1/2}$
since $M$ is a $2\times 2$ positive-definite matrix. Next, we have that
\begin{equation*}
    \det M=\det(\Sigma_0^{1/2}) \det(\Sigma_t) \det(\Sigma_0^{1/2}) = \det \Sigma_0 \cdot \det \Sigma_t=(\det \Sigma_0)^2=\frac{a^4b^4}{256},
\end{equation*}
where we used the similarity of $\Sigma_t$ and $\Sigma_0$.
Furthermore, using the cyclical invariance of the trace we have that
\begin{equation*}
\begin{split}
    \operatorname{tr} M=& \operatorname{tr} \left( \Sigma_0^{1/2} \Sigma_t \Sigma_0^{1/2} \right)=\operatorname{tr} \left(  \Sigma_t \Sigma_0^{1/2} \Sigma_0^{1/2}\right)=\operatorname{tr}(\Sigma_t \Sigma_0)\\
    =&\operatorname{tr}\left[\begin{pmatrix}
    \frac{a^2 \cos^2 t+b^2 \sin^2 t}{4}& \frac{(a^2-b^2)\cos t \sin t}{4}\\
    \frac{(a^2-b^2)\cos t \sin t}{4} & \frac{a^2 \sin^2 t+b^2 \cos^2 t}{4}
    \end{pmatrix}\cdot \begin{pmatrix}
    \frac{a^2}{4}& 0\\
    0 & \frac{b^2}{4}
    \end{pmatrix} \right]\\
    =&\frac{(a^4+b^4)\cos^2 t +2a^2 b^2 \sin^2 t}{16}.
\end{split}
\end{equation*}
Hence,
\begin{equation*}
    \begin{split}
        \operatorname{tr}(M^{1/2})=&\left(\operatorname{tr}(M)+2 \sqrt{\det(M)} \right)^{1/2}\\
        =&\left(\frac{(a^4+b^4)\cos^2 t +2a^2 b^2 \sin^2 t}{16}+2\sqrt{\frac{a^4b^4}{256}}\right)^{1/2}  \\
        =& \left(\frac{(a^4+b^4)\cos^2 t +2a^2 b^2 \sin^2 t+2a^2b^2}{16} \right)^{1/2} \\
        =& \frac{\sqrt{(a^2+b^2)^2\cos^2 t +4a^2 b^2 \sin^2 t} }{4}.
    \end{split}
\end{equation*}
Finally, applying the similarity of $\Sigma_t$ and $\Sigma_0$ again we have that
    $\operatorname{tr} \Sigma_t=\operatorname{tr} \Sigma_0=\frac{a^2+b^2}{4},$
and so
\begin{equation*}
\begin{split}
    W_2^2(\mu_0,\mu_t)=&|m_0-m_t|^2+\operatorname{tr}\left(\Sigma_0+\Sigma_t-2\left(\Sigma_0^{1/2}\Sigma_t \Sigma_0^{1/2}\right)^{1/2}\right)\\
    =&|u-R_t u|^2+2\operatorname{tr} \Sigma_0-2 \operatorname{tr}(M^{1/2})\\
    =&4|u|^2\sin^2 \frac{t}{2}+\frac{a^2+b^2-\sqrt{(a^2+b^2)^2\cos^2 t +4a^2 b^2 \sin^2 t}}{2},
\end{split}
\end{equation*}
which concludes the proof.
\end{proof}

We can now prove that rotation manifolds are not isometric to circles in general.

\begin{theorem}\label{thm:w-non-iso}
Assume that $a,b>0$, and $\mu_0$ is the uniform probability measure over the elliptical domain
\begin{equation*}
\mathcal{E}_0=\left\{(x_1,x_2)\in \R^2~:~\frac{x_1^2}{a^2}+\frac{x_2^2}{b^2} \leq 1\right\}.
\end{equation*}
Furthermore, assume that $\mu_t=(R_{t}x)\sharp \mu_0$. Then $\left(\{\mu_t\}_{t\in [0,2\pi]},W_2\right)$ is isometric to a circle if and only if $a=b$.
\end{theorem}
\begin{proof}
When $a=b$ we have that $\mu_t=\mu_0$ for all $t\in [0,2\pi]$, so $\left(\{\mu_t\}_{t\in [0,2\pi]},W_2\right)$ is isometric to a (degenerate) circle. 

Now assume that $a\neq b$ and suppose by contradiction that there exists $\gamma:[0,2\pi] \to \partial B_r$ for some $r>0$ such that
\begin{equation*}
    W_2^2(\mu_s,\mu_t)=|\gamma(s)-\gamma(t)|^2,\quad \forall s,t \in [0,2\pi].
\end{equation*}
From Proposition~\ref{prp:analytic_distances} we have that
\begin{equation*}
    |\gamma(0)-\gamma(\pi/4)|=|\gamma(\pi/4)-\gamma(\pi/2)|=|\gamma(\pi/2)-\gamma(3\pi/4)|=|\gamma(3\pi/4)-\gamma(\pi)|.
\end{equation*}
Additionally, we have that $|\gamma(0)-\gamma(\pi)|=0$, and so $\gamma(0)=\gamma(\pi)$. This means that the quadrilateral with vertices $\gamma(0),\gamma(\pi/4),\gamma(\pi/2),\gamma(3\pi/2)$ is a rhombus. Since the only rhombus that can be inscribed in a circle is the square, we obtain that $\gamma(0),\gamma(\pi/4),\gamma(\pi/2),\gamma(3\pi/2)$ form a square. The latter means that
\begin{equation*}
    |\gamma(\pi/2)-\gamma(0)|^2= 2|\gamma(\pi/4)-\gamma(0)|^2.
\end{equation*}
From Proposition~\ref{prp:analytic_distances} we have that
\begin{equation*}
\begin{split}
    |\gamma(\pi/2)-\gamma(0)|^2=&\frac{(a-b)^2}{2},\\
    |\gamma(\pi/4)-\gamma(0)|^2=&\frac{a^2+b^2-\sqrt{\frac{(a^2+b^2)^2}{2}+2a^2b^2}}{2},
\end{split}
\end{equation*}
and so we obtain
\begin{equation*}
    \frac{(a-b)^2}{2}=a^2+b^2-\sqrt{\frac{(a^2+b^2)^2}{2}+2a^2b^2}.
\end{equation*}
Simplifying the latter yields
    $(a-b)^4=0,$
which is a contradiction.
\end{proof}

We can also prove a more general version of Theorem~\ref{thm:w-non-iso} with elliptical domains not necessarily centered at the origin.
\begin{theorem}\label{thm:w-non-iso-u}
Assume that $a,b>0$, and $\mu_0$ is the uniform probability measure over the elliptical domain
\begin{equation*}
\mathcal{E}=\left\{(x_1,x_2)\in \R^2~:~\frac{(x_1-u_1)^2}{a^2}+\frac{(x_2-u_2)^2}{b^2} \leq 1\right\}.
\end{equation*}
Furthermore, assume that $\mu_t=(R_{t}x)\sharp \mu_0$. Then $\left(\{\mu_t\}_{t\in [0,2\pi]},W_2\right)$ is isometric to a circle if and only if $a=b$.
\end{theorem}
\begin{proof}
Denote by $u=(u_1,u_2)$. Assume that $u\neq 0$ for otherwise Theorem~\ref{thm:w-non-iso} applies. If $a=b$ then from~\eqref{WssTrue} we have that
\begin{equation*}
    W_2^2(\mu_s,\mu_t)=4|u|^2\sin^2 \frac{t-s}{2}=|R_s u - R_t u|^2,
\end{equation*}
which means that $\mu_t \mapsto R_t u$ is an isometry; that is, $\left(\{\mu_t\}_{t\in [0,2\pi]},W_2\right)$ is isometric to the circle of radius $|u|$.

Now assume that $a\neq b$, and suppose by contradiction that there exists $\gamma:[0,2\pi] \to \partial B_r$ for some $r>0$ such that
\begin{equation*}
    W_2^2(\mu_s,\mu_t)=|\gamma(s)-\gamma(t)|^2,\quad \forall s,t \in [0,2\pi].
\end{equation*}
Similar to the proof of Theorem~\ref{thm:w-non-iso} we have that $\gamma(0),\gamma(\pi/2),\gamma(\pi),\gamma(3\pi/2)$ are vertices of a square. Indeed, the only difference is that $u\neq0$ implies that $\{\mu_t\}$ and $\{\gamma(t)\}$ are $2\pi$-periodic instead of $\pi$-periodic. Hence, we must have
\begin{equation*}
    |\gamma(\pi)-\gamma(0)|^2=2|\gamma(\pi/2)-\gamma(0)|^2.
\end{equation*}
Invoking Proposition~\ref{prp:analytic_distances} we find that
\begin{equation*}
\begin{split}
    |\gamma(\pi)-\gamma(0)|^2=&4|u|^2,\\
    |\gamma(\pi/2)-\gamma(0)|^2=&2|u|^2+\frac{(a-b)^2}{2},
\end{split}
\end{equation*}
and so we obtain
    $4|u|^2=4|u|^2+(a-b)^2,$
which is a contradiction.
\end{proof}

\begin{remark}\label{rmk:gen-ellipse}
Proposition~\ref{prp:analytic_distances} and Theorems~\ref{thm:w-non-iso}, \ref{thm:w-non-iso-u} can be of course extended to uniform measures over general elliptical domains
\begin{equation*}
    \mathcal{E}=\left\{x\in \mathbb{R}^2~:~ (x-u)^\top \Pi^{-1}(x-u)  \leq 1 \right\},
\end{equation*}
where $u\in \mathbb{R}^2$, and $\Pi$ is symmetric and positive definite. Indeed, if $\mu_0$ is the uniform probability measure over $E$, then $\mu_{t_0}=(R_{t_0}x)\sharp \mu_0$ is the uniform probability measure over
\begin{equation*}
    E'=\left\{(x_1,x_2)\in \mathbb{R}^2~:~\frac{(x_1-(R_{t_0}u)_1)^2}{\lambda_1}+\frac{(x_2-(R_{t_0}u)_2)^2}{\lambda_2} \leq 1 \right\},
\end{equation*}
where $\lambda_1,\lambda_2>0$ are the eigenvalues of $\Pi$. Hence, we have that
\begin{equation*}
    \begin{split}
        &W_2^2(\mu_s,\mu_t)\\
        =&W_2^2(\mu_{t_0},\mu_{t_0+t-s})=W_2^2(\mu_{t_0},(R_{t-s}x)\sharp\mu_{t_0})\\
        =&4|R_{t_0}u|^2 \sin^2 \frac{t-s}{2}+\frac{\lambda_1+\lambda_2-\sqrt{(\lambda_1+\lambda_2)^2\cos^2(t-s)+4\lambda_1 \lambda_2 \sin^2 (t-s)}}{2}\\
        =&4|u|^2 \sin^2 \frac{t-s}{2}+\frac{\operatorname{tr} \Pi-\sqrt{(\operatorname{tr}\Pi)^2\cos^2(t-s)+4\det \Pi ~\sin^2 (t-s)}}{2}.
    \end{split}
\end{equation*}
\end{remark}

\subsection{Linearized Wasserstein distances}

Let $\nu \in \mathcal{P}_{ac,2}(\mathbb{R}^2)$. Recall that for $\mu_1,\mu_2 \in \mathcal{P}_2(\mathbb{R}^2)$ we define the linearized Wasserstein distance with a reference measure $\nu$ as
\begin{equation*}
    W_{2,\nu}(\mu_1,\mu_2)=\left( \int_{\mathbb{R}^2}|T_2(x)-T_1(x)|^2 d\nu(x)\right)^{\frac{1}{2}},
\end{equation*}
where $T_1,T_2$ are optimal transportation maps pushing $\nu$ forward to $\mu_1,\mu_2$, respectively.

\begin{lemma}\label{lma:rotation_LOT}
Assume that $\nu \in \mathcal{P}_{ac,2}(\mathbb{R}^2)$ is rotationally symmetric; that is, $(R_{t}x)\sharp \nu = \nu$ for all $t\in \mathbb{R}$. Furthermore, assume that $\mu_t=(R_{t}x)\sharp \mu_0$ for some $\mu_0 \in \mathcal{P}_2(\mathbb{R}^2)$, and denote by $T_t$ the optimal transportation map that pushes $\nu$ forward to $\mu_t$. Then we have that
\begin{equation*}
    T_t(x)=R_{t}T_0(R_{-t}x),~ \text{a.e. in}~ \mathbb{R}^2,~0\leq t \leq 2\pi,
\end{equation*}
and
\begin{equation*}
    W_{2,\nu}(\mu_s,\mu_t)=W_{2,\nu}(\mu_0,\mu_{t-s}),\quad 0\leq s<t \leq 2\pi.
\end{equation*}
\end{lemma}
\begin{proof}
Since $T_0$ is the optimal transportation map between $\nu$ and $\mu_0$ we have that $\pi_0=(x,T_0(x))\sharp \nu$ is the optimal transportation plan between $\nu$ and $\mu_0$. Applying the arguments in Lemma~\ref{lma:rotation} we have that $\pi_t=(R_{t}x,R_{t}y)\sharp \pi_0$ is the optimal transportation plan between $(R_{t}x)\sharp \nu=\nu$ and $(R_{t}x)\sharp \mu_0=\mu_t$. Note that
\begin{equation*}
    \pi_t=(R_{t}x,R_{t}y)\sharp \pi_0=\pi_t=(R_{t}x,R_{t}T_0(x))\sharp \nu.
\end{equation*}
Furthermore, taking into account that $\nu=(R_{-t}x)\sharp \nu$ we obtain that
\begin{equation*}
    \pi_t=(R_{t}x,R_{t}T_0(x))\sharp \nu=(x,R_{t}T_0(R_{-t}x))\sharp \nu,
\end{equation*}
and so $\pi_t$ is generated by a necessarily optimal transportation map $x\mapsto R_{t} T_0(R_{-t}x)$. But optimal transportation map from $\nu$ to $\mu_t$ is unique by Brenier's theorem, and so
\begin{equation*}
    T_t(x)=R_{t} T_0(R_{-t}x),\quad \text{a.e. in}~\mathbb{R}^2.
\end{equation*}
Next, we have that
\begin{equation*}
    \begin{split}
        &W_{2,\nu}^2(\mu_s,\mu_t)\\
        =&\int_{\mathbb{R}^2} |T_t(x)-T_s(x)|^2 d\nu(x)=\int_{\mathbb{R}^2} |R_{t}T_0(R_{-t}x)-R_sT_0(R_{-s}x)|^2 d\nu(x)\\
    =&\int_{\mathbb{R}^2} |R_{t-s}T_0(R_{-t}x)-T_0(R_{-s}x)|^2 d\nu(x)\\
    =&\int_{\mathbb{R}^2} |R_{t-s}T_0(R_{s-t}x)-T_0(x)|^2 d\nu(x)=\int_{\mathbb{R}^2} |T_{t-s}(x)-T_0(x)|^2 d\nu(x)\\
    &=W_{2,\nu}^2(\mu_0,\mu_{t-s}),
    \end{split}
\end{equation*}
which concludes the proof.
\end{proof}
 
\begin{proposition}\label{prp:analytic-distances-LOT}
Assume that $\nu$ is a standard normal distribution, and $\mu_0$ is a Gaussian with mean $u\in \mathbb{R}^2$ and covariance matrix $\Sigma_0=\begin{pmatrix} a^2&0\\0&b^2 \end{pmatrix}$ for some $a,b>0$. Furthermore, let $\mu_t=(R_t x)\sharp \mu_0$. Then we have that 
\begin{equation}\label{LOT:WssTrue}
    W_{2,\nu}^2(\mu_s,\mu_t)=4|u|^2 \sin^2 \frac{t-s}{2}+2(a-b)^2 \sin^2 (t-s),\quad 0\leq s<t \leq 2\pi.
\end{equation}
\end{proposition}
\begin{proof}
Taking into account Lemma~\ref{lma:rotation_LOT} it is enough to prove~\eqref{LOT:WssTrue} for $s=0$. Furthermore, we have that $\mu_t=(R_t x)\sharp \mu_0$ is a Gaussian with mean $R_t u$ and covariance matrix $\Sigma_t=R_t \Sigma_0 R_{-t}$. Hence, from~\cite{gelbrich90} (see also~\cite[Remarks 2.31-32]{peyre19computational}) we have that
\begin{equation*}
    T_t(x)=R_t u+\Sigma_t^{\frac{1}{2}} x,\quad x\in \mathbb{R}^2.
\end{equation*}
Furthermore, recalling that
\begin{equation*}
    \int_{\mathbb{R}^2} |m+Mx|^2 d\nu(x)=\operatorname{tr}(M^\top M)+|m|^2,
\end{equation*}
we obtain
\begin{equation*}
    \begin{split}
        &W_{2,\nu}^2(\mu_0,\mu_t)\\=&\int_{\mathbb{R}^2}|T_t(x)-T_0(x)|^2 d\nu(x)=\int_{\mathbb{R}^2}\left|R_t(u)-u+(\Sigma_t^{\frac{1}{2}}-\Sigma_0^{\frac{1}{2}}x)\right|^2 d\nu(x)\\
        =&|R_tu -u|^2+\operatorname{tr}(\Sigma_t^{\frac{1}{2}}-\Sigma_0^{\frac{1}{2}})^2=4|u|^2 \sin^2 \frac{t}{2}+\operatorname{tr}\left(\Sigma_t+\Sigma_0-\Sigma_t^{\frac{1}{2}} \Sigma_0^{\frac{1}{2}}-\Sigma_0^{\frac{1}{2}} \Sigma_t^{\frac{1}{2}} \right)\\
        =& 4|u|^2 \sin^2 \frac{t}{2}+\operatorname{tr} \Sigma_t+\operatorname{tr} \Sigma_0-2\operatorname{tr}\left( \Sigma_t^{\frac{1}{2}} \Sigma_0^{\frac{1}{2}}\right)\\
        =&4|u|^2 \sin^2 \frac{t}{2}+2\operatorname{tr} \Sigma_0-2\operatorname{tr}\left( \Sigma_t^{\frac{1}{2}} \Sigma_0^{\frac{1}{2}}\right),
    \end{split}
\end{equation*}
where we used the similarity of $\Sigma_t$ and $\Sigma_0$, and the cyclic invariance of the trace. Next, we have that
\begin{equation*}
    \Sigma_t^{\frac{1}{2}}=\left( R_t \Sigma_0 R_{-t}\right)^{\frac{1}{2}}= R_t \Sigma_0^{\frac{1}{2}} R_{-t}=\begin{pmatrix} a \cos^2 t+b \sin^2 t& (a-b)\sin t \cos t\\ (a-b)\sin t \cos t& a \sin^2 t+b \cos^2 t  \end{pmatrix},
\end{equation*}
and so
\begin{equation*}
\begin{split}
    \operatorname{tr} \left( \Sigma_t^{\frac{1}{2}} \Sigma_0^{\frac{1}{2}}\right)=&\operatorname{tr} \left[\begin{pmatrix} a \cos^2 t+b \sin^2 t& (a-b)\sin t \cos t\\ (a-b)\sin t \cos t& a \sin^2 t+b \cos^2 t  \end{pmatrix} \cdot \begin{pmatrix} a & 0\\ 0& b  \end{pmatrix}\right]\\
    =&(a^2+b^2)\cos^2 t+2ab \sin^2 t.
\end{split}
\end{equation*}
Therefore we have that
\begin{equation*}
    \begin{split}
        W_{2,\nu}^2(\mu_0,\mu_t)=&4|u|^2\sin^2 \frac{t}{2}+2(a^2+b^2)-2(a^2+b^2)\cos^2 t-4ab\sin^2 t\\
        =&4|u|^2 \sin^2 \frac{t}{2}+2(a-b)^2 \sin^2 t,
    \end{split}
\end{equation*}
which concludes the proof.
\end{proof}

\begin{theorem}\label{thm:LOT-non-iso}
Assume that $\nu$ is a standard normal distribution, and $\mu_0$ is a Gaussian with mean $u\in \mathbb{R}^2$ and covariance matrix $\Sigma_0=\begin{pmatrix} a^2& 0 \\ 0 & b^2 \end{pmatrix}$ for some $a,b>0$. Furthermore, let $\mu_t=(R_t x)\sharp \mu_0$. Then $\left(\left\{\mu_t\right\}_{t\in [0,2\pi]},W_{2,\nu} \right)$ is isometric to a circle if and only if $u=0$ or $a=b$.
\end{theorem}
\begin{proof}
If $u=0$ then~\eqref{LOT:WssTrue} yields that

\begin{equation*}
    W_{2,\nu}(\mu_s,\mu_t)^2=2(a-b)^2 \sin^2 (t-s) = \frac{(a-b)^2}{2}|R_{2s}e_1-R_{2t}e_1|^2,
\end{equation*}

which means that $\mu_t \mapsto \frac{|a-b|}{\sqrt{2}} R_{2t}e_1$ is an isometry, where $e_1=(1,0)$.

If $a=b$ then~\eqref{LOT:WssTrue} yields that

\begin{equation*}
    W_{2,\nu}(\mu_s,\mu_t)=4|u|^2\sin^2 \frac{t-s}{2} = |R_s u-R_t u|^2,
\end{equation*}

which means that $\mu_t \mapsto R_t u$ is an isometry.

Now assume that $u\neq 0$, $a\neq b$, and suppose by contradiction that there exists $\gamma:[0,2\pi]\to \partial B_r$ for some $r>0$ such that
\begin{equation*}
    W_{2,\nu}^2(\mu_s,\mu_t)=|\gamma(s)-\gamma(t)|^2,\quad \forall s,t \in [0,2\pi].
\end{equation*}
Similar to Theorems~\ref{thm:w-non-iso} and~\ref{thm:w-non-iso-u} we have that $\gamma(0),\gamma(\pi/2),\gamma(\pi),\gamma(3\pi/2)$ are vertices of a square, and so
    $|\gamma(\pi)-\gamma(0)|^2=2|\gamma(\pi/2)-\gamma(0)|^2.$\\
From \eqref{LOT:WssTrue} we have that
\begin{equation*}
    \begin{split}
        |\gamma(\pi)-\gamma(0)|^2=&4|u|^2,\\
        |\gamma(\pi/2)-\gamma(0)|^2=&2|u|^2+2(a-b)^2,
    \end{split}
\end{equation*}
and so we obtain
    $4|u|^2=4|u|^2+4(a-b)^2,$
which is a contradiction.
\end{proof}

\begin{remark}\label{rmk:gen}
As in Remark~\ref{rmk:gen-ellipse}, we note that Proposition~\ref{prp:analytic-distances-LOT} and Theorem~\ref{thm:LOT-non-iso} naturally extend to Gaussians $\mu_0$ with a generic covariance matrix $\Sigma_0$. Indeed, for a suitable angle $t_0$ we have that $\mu_{t_0}$ is a Gaussian with mean $m_{t_0}=R_{t_0}u$ and covariance $\Sigma_{t_0}=\begin{pmatrix} \lambda_1 & 0\\ 0& \lambda_2\end{pmatrix}$, where $\lambda_1,\lambda_2>0$ are the eigenvalues of $\Sigma_0$. But then we have that

\begin{equation*}
    \begin{split}
    W_{2,\nu}^2(\mu_s,\mu_t)=&W_{2,\nu}^2(\mu_{t_0},\mu_{t_0+t-s})=W_{2,\nu}^2(\mu_{t_0},(R_{t-s}x)\sharp\mu_{t_0})\\
    =&4|R_{t_0}u|^2 \sin^2 \frac{t-s}{2}+2(\sqrt{\lambda_1}-\sqrt{\lambda_2})^2 \sin^2(t-s)\\
    =&4|u|^2 \sin^2 \frac{t-s}{2}+2(\lambda_1+\lambda_2-2\sqrt{\lambda_1 \lambda_2}) \sin^2(t-s)\\
    =&4|u|^2 \sin^2 \frac{t-s}{2}+2(\operatorname{tr} \Sigma_0-2\sqrt{\det \Sigma_0}) \sin^2(t-s).
    \end{split}
\end{equation*}

\end{remark}

\subsection{No-collision distances}

Here, we show that no-collision distances in general do not provide an isometry between rotated measures and planar circles either.

\begin{theorem}\label{thm:collision-non-iso}
Assume that $\mu_0$ is a uniform measure over an elliptical domain
\begin{equation*}
    \mathcal{E}=\left\{(x_1,x_2) \in \mathbb{R}^2~:~\frac{(x_1-u_1)^2}{a^2}+\frac{(x_2-u_2)^2}{b^2}\leq 1 \right\},
\end{equation*}
where $u=(u_1,u_2)\neq 0$, and $a,b>0$. Furthermore, assume that $\mu_t=(R_tx)\sharp \mu_0$, and $\mathcal{S}$ is a slicing schedule such as in Theorem~\ref{thm:dilation_isometry}. Then $\left(\{\mu_t\}_{t\in[0,2\pi]},W_{\mathcal{S},2}\right)$ is isometric to a circle if and only if $a=b$.
\end{theorem}
\begin{proof}
When $a=b$ we have that $\mu_t$ is the uniform measure over the disk of radius $a$ centered at $R_t u$. Therefore, $\mu_t=(R_t u-u+x)\sharp \mu_0$, and by Theorem~\ref{thm:translation_isometry} we have that
\begin{equation*}
    W_{\mathcal{S},2}(\mu_s,\mu_t)=|R_s u - R_t u|,\quad \forall s,t,
\end{equation*}
which means that $\mu_t \mapsto R_t u$ is an isometry.

Now assume that $a\neq b$, and suppose by contradiction that there exists $\gamma:[0,2\pi]\to \partial B_r$ for some $r>0$ such that
\begin{equation*}
    W_{\mathcal{S},2}^2(\mu_s,\mu_t)=|\gamma(s)-\gamma(t)|^2,\quad \forall s,t \in [0,2\pi].
\end{equation*}

We have that $\mu_{\pi/2}$ is the uniform measure over the elliptical domain
\begin{equation*}
    \mathcal{E}'=\left\{(x_1,x_2) \in \mathbb{R}^2~:~\frac{(x_1+u_2)^2}{b^2}+\frac{(x_2-u_1)^2}{a^2}\leq 1 \right\}.
\end{equation*}
Furthermore, denote by $\nu,\nu'$ respectively the uniform measures over elliptical domains
\begin{equation*}
\mathcal{E}_0=\left\{(x_1,x_2) \in \mathbb{R}^2~:~\frac{x_1^2}{a^2}+\frac{x_2^2}{b^2}\leq 1 \right\},\quad  \mathcal{E}'_0=\left\{(x_1,x_2) \in \mathbb{R}^2~:~\frac{x_1^2}{b^2}+\frac{x_2^2}{a^2}\leq 1 \right\},
\end{equation*}
From Theorems~\ref{thm:translation_isometry} and~\ref{thm:dilation_isometry} we have the following no-collision transportation maps
\begin{equation*}
    \begin{split}
        t_{\mu_0}^\nu(x_1,x_2)=&(x_1-u_1,x_2-u_2),\\
        t_{\nu}^{\nu'}(x_1,x_2)=&\left(\frac{b}{a}x_1,\frac{a}{b}x_2 \right),\\
        t_{\nu'}^{\mu_{\pi/2}}(x_1,x_2)=&(x_1-u_2,x_2+u_1).
    \end{split}
\end{equation*}
Hence, by~\cite[Proposition 2]{nurbekyan20nocollision}, which follows from Part 1 in Theorem~\ref{thm:maps_existence}, we have that
\begin{equation*}
    t_{\mu_0}^{\mu_{\pi/2}}(x_1,x_2)=t_{\nu'}^{\mu_{\pi/2}}\circ t_{\nu}^{\nu'} \circ t_{\mu_0}^\nu(x_1,x_2)=\left( \frac{b}{a}(x_1-u_1)-u_2,\frac{a}{b}(x_2-u_2)+u_1 \right)
\end{equation*}
is the no-collision map pushing forward $\mu_0$ to $\mu_{\pi/2}$. Note that $t_{\mu_0}^{\mu_{\pi/2}}=\nabla \phi$, where
\begin{equation*}
    \phi(x_1,x_2)=\frac{b(x_1-u_1)^2}{2a}-u_2x_1+\frac{a(x_2-u_2)^2}{2b}+u_1x_2.
\end{equation*}
Since $\phi$ is convex, Brenier's theorem yields that $t_{\mu_0}^{\mu_{\pi/2}}$ is the optimal transportation map pushing $\mu_0$ optimally to $\mu_{\pi/2}$, and so
$W_{\mathcal{S},2}(\mu_0,\mu_{\pi/2})=W_2(\mu_0,\mu_{\pi/2}).$\\
Similarly, we can prove that
$W_{\mathcal{S},2}(\mu_{\pi k/2},\mu_{\pi l/2})=W_2(\mu_{\pi k/2},\mu_{\pi l/2}),\; \forall k,l \in \mathbb{Z}.$\\
From Proposition~\ref{prp:analytic_distances} we have that
\begin{equation*}
W_2(\mu_0,\mu_{\pi/2})=W_2(\mu_{\pi/2},\mu_{\pi})=W_2(\mu_\pi,\mu_{3\pi/2})=W_2(\mu_{3\pi/2},\mu_{0}),
\end{equation*}
and so
$W_{\mathcal{S},2}(\mu_0,\mu_{\pi/2})=W_{\mathcal{S},2}(\mu_{\pi/2},\mu_{\pi})=W_{\mathcal{S},2}(\mu_\pi,\mu_{3\pi/2})=W_{\mathcal{S},2}(\mu_{3\pi/2},\mu_{0}).$\\
The latter yields
\begin{equation*}
    |\gamma(0)-\gamma(\pi/2)|=|\gamma(\pi/2)-\gamma(\pi)|=|\gamma(\pi)-\gamma(3\pi/2)|=|\gamma(3\pi/2)-\gamma(0)|,
\end{equation*}
by the isometry assumption. Thus, $\gamma(0),\gamma(\pi/2),\gamma(\pi),\gamma(3\pi/2)$ are vertices of a rhombus that must necessarily be a square since $\{\gamma(t)\}\subset \partial B_r$. \\Therefore we have that
    $|\gamma(0)-\gamma(\pi)|^2=2|\gamma(0)-\gamma(\pi/2)|^2.$ 
But
\begin{equation*}
    \begin{split}
        |\gamma(0)-\gamma(\pi)|^2=&W_{\mathcal{S},2}^2(\mu_0,\mu_\pi)=W_2^2(\mu_0,\mu_\pi),\\
        |\gamma(0)-\gamma(\pi/2)|^2=&W_{\mathcal{S},2}^2(\mu_0,\mu_{\pi/2})=W_2^2(\mu_0,\mu_{\pi/2}),
    \end{split}
\end{equation*}
and we arrive at a contradiction by repeating the calculation in Theorem~\ref{thm:w-non-iso-u}.
\end{proof}

\section{Experimental results}\label{Exper}

In this section, we provide several experiments \footnote{Code for this work is available at \href{https://github.com/enegrini/Applications-of-No-Collision-Transportation-Maps-in-Manifold-Learning.git}{github.com/enegrini/Applications-of-No-Collision}.} that demonstrate our theoretical results. In particular we perform manifold learning on synthetically generated $2$-dimensional images using no-collision distances and compare our results with other manifold learning techniques. For more examples we refer to the Supplementary Materials.

For our experiments we generate synthetic data as follows. We fix $\mu_0 \in \mathcal{P}_{ac}(\mathbb{R}^d)$ and consider
\begin{equation*}
 \mathcal{M}(\mu_0,\Theta)=\left\{ F_\theta \sharp \mu_0~:~\theta \in \Theta\right\},
\end{equation*}
where $F_\theta$ is a parametric mapping. In particular, we have
\begin{itemize}
    \item $F_\theta(x)=x-\theta$ for translations,
    \item $F_\theta(x)=\theta \odot x$ for dilations,
    \item $F_\theta(x)=R_\theta x$ for for rotations.
\end{itemize}
We then sample $\{\theta_1, \cdots,\theta_m\} \subset \Theta $ and consider
\begin{equation*}
    \left\{\mu_1,\mu_2,\cdots,\mu_m\right\}=\left\{F_{\theta_1}\sharp \mu_0,F_{\theta_2}\sharp \mu_0,\cdots,F_{\theta_m}\sharp \mu\right\}.
\end{equation*}
In practice, we work with discrete versions of these distributions: we represent them discretely as an $n\times n$ pixel image. Each point in the support of the image has a pixel value which (after normalization) is the mass associated with the density $\mu_i$. In the following experiments, $n$, the grid size for the discrete distribution, is fixed at $n = 128$. In practice, the choice of $n$ will influence the quality of the embedding since a coarser grid may result in an uneven no-collision splitting and consequently in a lower quality embedding. Examples for different choices of $n$ can be found in the supplementary material.

Once the data is generated, the embedding is performed as follows: given the set of observed transformed distributions, we compute the distance matrix containing all the pairwise distances. We then perform MDS on the distance matrix to find a $2$-dimensional embedding. In each example we compute the Euclidean distance matrix, Wasserstein distance matrix, LOT distance matrix (using the Python Optimal Transport (POT) package \cite{JMLR:v22:20-451}), and no-collision distance matrices. 
 Another possibility is to perform the embedding following the approach proposed in  \cite{cloninger23linearized}. In this work the authors propose an algorithm based on SVD which is applied directly on features, without the need of computing the distance matrix. In this work we chose to use MDS since Wassmap does not provide features so the SVD based approach could not be used. However, a comparison of LOT and no-collision embedding using the approach proposed in \cite{cloninger23linearized} can be found in Section \ref{SVDEmb} in the Supplementary Material.

Specifically, we compute the no-collision distance matrix applying Algorithm~\ref{alg:no-col} to each of the observed distributions for generating corresponding no-collision features (geometric centers or centers of mass) and computing pairwise Euclidean distances between these features. For each example, we assume that images have the same size and we choose the number of cuts, $N$, that gives the best visual results. In particular, we start with $N=2$ and increase the number of cuts until the result start decreasing in accuracy. For large enough $N$ the problem reduces to performing MDS on the original pixel images, and experiments show that this case provides inaccurate manifold reconstructions.

We compare our results for manifold learning with MDS, Isomap, and Diffusion Maps on the original pixel images using Euclidean distances; with Wassmap; and with MDS on LOT features. For Wassmap and LOT, we generate pointcloud data from the images as done in the original papers. For LOT, we pick the reference distribution to be a Gaussian with given mean and covariance matrix. For Isomap, the $k$-NN graph is used to estimate geodesics and $k$ is picked to give the best visual results. Finally, for Diffusion maps we use a Gaussian kernel and select the parameter epsilon according to the algorithm proposed in \cite{berry2015nonparametric}.

\begin{remark}\label{rescale}
    In this work we are mainly interested in recovering correctly only the relative $W_2$ distance between data points, that is, the true $W_2$ distance up to a scaling factor. This is usually sufficient for most application such as clustering or classification, however for manifold learning application it is interesting to also recover the correct scale. We propose a simple and computationally inexpensive way of doing so. No-collision maps produce a distance matrix that differ from the true $W_2$ distance matrix only by a scaling factor. As a consequence, in order to re-scale the no-collision distance matrix correctly one only needs to enforce that one of such distances coincides with the true $W_2$ distance. In practice, we re-scale the no-collision distance matrix by enforcing that the largest no-collision distance coincides with the largest $W_2$ distance. Note that this approach does not have a large impact on the computational time since it only requires one $W_2$ distance computation. In the following experiments we corrected the scale of no-collision distance matrices using the strategy above and are able to recover the correct scaling as well as the correct embedding.\\
    A second possibility is to keep track of all the constants and scalings which we originally dropped in our computations. Moreover, since our algorithm is implemented for integer grids, where coordinates of pixels coincide with pixel indices, we also have to rescale our image data to match the physical bounds used in the Wasserstein distance computations. In other words, in our current implementation the scale of no-collision features (coordinates of center of mass or geometric center) depend on the image resolution, so we have to account for this when rescaling. Accounting for these additional constants also provides approximately accurate scaling, see for instance Supplementary Material Section \ref{SVDEmb}
\end{remark}

\subsection{Translation Manifold}
 We fix a base measure $\mu_0$ to be the indicator function of a disc of radius 1 centered at the origin. For a given translation set $\Theta = \left\{\theta=(\theta_1,\theta_2) \right\} \subset \mathbb{R}^2$ the corresponding translation manifold is given by:
$$\mathcal{M}^{\text{trans}}(\mu_0, \Theta) := \{ (x - \theta)\sharp\mu_0;\; \theta\in \Theta\}$$
In this example we consider a translation set $\Theta$ representing a triangular translation grid with 24 translation points. More examples with different translation grids can be found in the Supplementary Materials.

\begin{figure}[H]
    \centering
    \includegraphics[width = \linewidth]{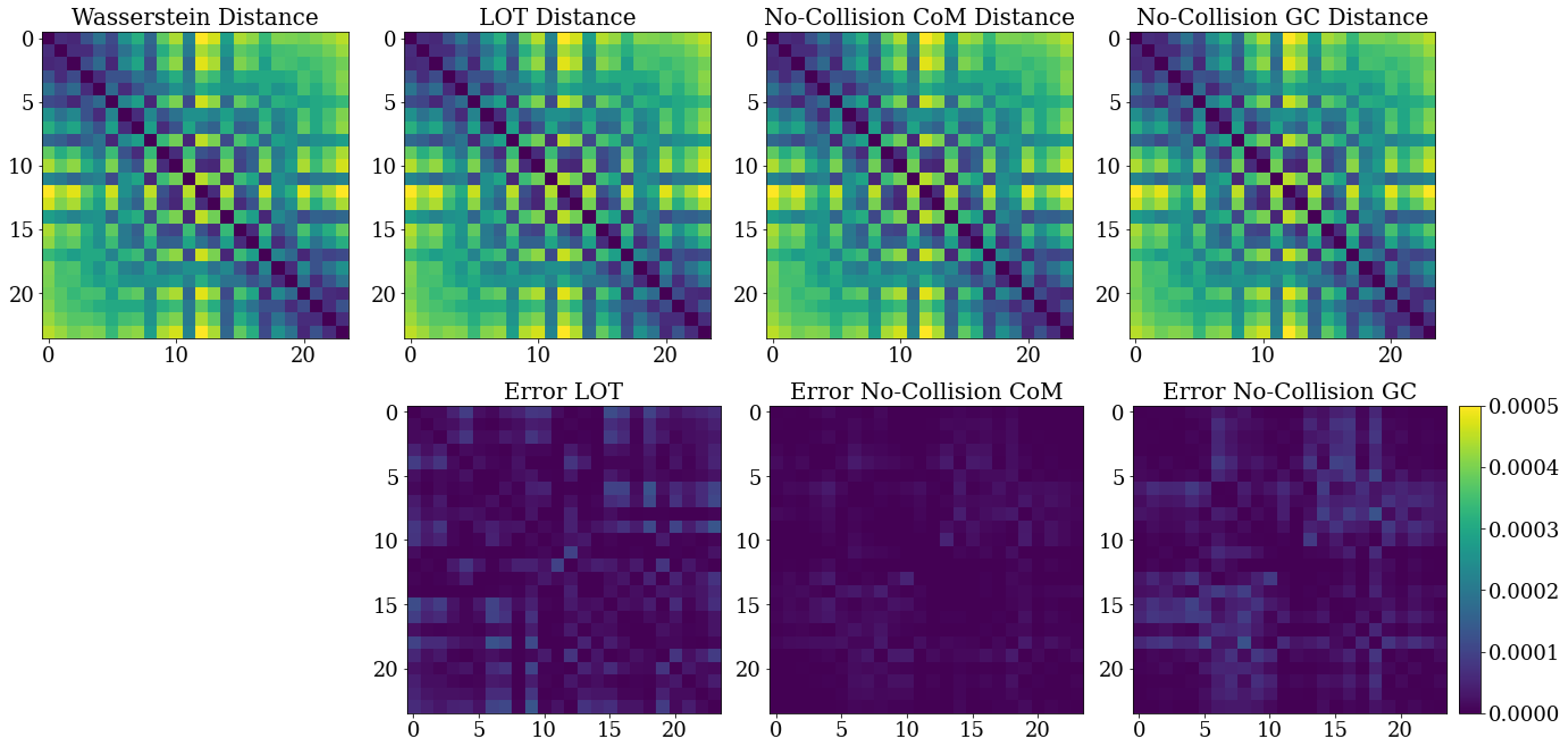}
    \caption{Distance matrix comparison for translation example on a triangular grid. \textbf{Top:} Distance matrices. \textbf{Bottom:} Squared Error between approximate distance matrices and Wasserstein distance.}
    \label{Transl3Dist}
\end{figure}

Since both no-collision maps and LOT give an approximation of the true Wasserstein distance, the first question we address is how good these approximations are. In Figure \ref{Transl3Dist} we compare the distance matrices (properly re-scaled as explained in Remark \ref{rescale}) obtained by the different methods. In the first row, from left to right we represent the true Wasserstein distance, LOT distance and no-collision distances; in the second row we represent the $L^2$ error in the approximation of the Wasserstein distance. In this case for no-collision maps we use $N=2$ cuts since this was the choice that gave the best results both for center of mass and geometric center features. As reference for LOT in this case we picked one Gaussian centered at $(0,0)$ with covariance matrix $25I$. Picking more than one references gives similar results in this case but requires more computational time. In Table \ref{tab:transEr3} we report the relative error in Frobenius norm for each distance. From these experiments we can see that LOT and no-collision provide an accurate approximation of the Wasserstein distance with relative errors that are less than $0.9\%$. However, the no-collision distance approximation using center of mass is the one that provides the best accuracy with relative error less than $0.5\%$.

\vspace{-12pt}
\begin{table}[H]
\caption{Relative error in Frobenius norm of Wasserstein distance approximation using LOT and no-collision distances for translation example on a triangular grid.}
    \label{tab:transEr3}
\centering
\resizebox{0.68\linewidth}{!}{%
\begin{tabular}{|c|c|c|c|}
\hline
\textbf{Distance}       & LOT    & \begin{tabular}[c]{@{}c@{}}No-Collision\\ Center or Mass\end{tabular} & \begin{tabular}[c]{@{}c@{}}No-Collision\\ Geometric Center\end{tabular} \\ \hline
\textbf{Relative Error} & 0.879\% & 0.493\%                                                                & 0.861\%                                                                  \\ \hline
\end{tabular}}
\end{table}

\begin{figure}
    \centering
    \includegraphics[width = 0.39\linewidth]{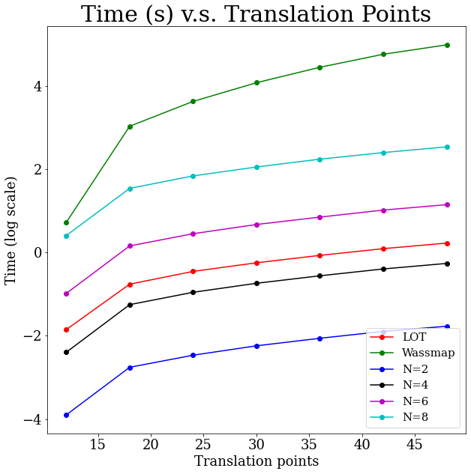}
    \caption{Computational time Wassmap, LOT and no-collision for multiple choices of $N$ and increasing number of translation points on a triangular grid.}
    \label{TransTimeTri}
\end{figure}

Not only no-collision distances give good approximations of the true Wasserstein distance, but they are also much faster to compute than the LOT and true Wasserstein distance since they do not require any optimization. Specifically, true Wasserstein distance computation took 36.9 seconds, LOT computation took 0.63 seconds while no-collision maps with $N=2$ cuts took 0.082 seconds which is 7.68 times faster than LOT and 450 times faster than Wassmap. To explore the computational time aspect more in depth we compared in Figure \ref{TransTimeTri} the time (in log scale) needed to obtain the distance matrix for Wassmap, LOT and no-collision for an increasing number of translation points. For no-collision $N=2$ already gave accurate results for all amounts of translation points, but for completeness we also report the time required for larger $N$ that also provided accurate distance approximations. We see that the computational time for Wasserstein distance is always larger than the one needed for LOT and no-collision and it also increases faster as the number of translation points increases. Comparing LOT and no-collision we see that LOT computational time is larger than no-collision when $N\leq4$, however no-collision already provides the best approximation for $N=2$.

\begin{figure}
    \centering
    \includegraphics[width = \linewidth]{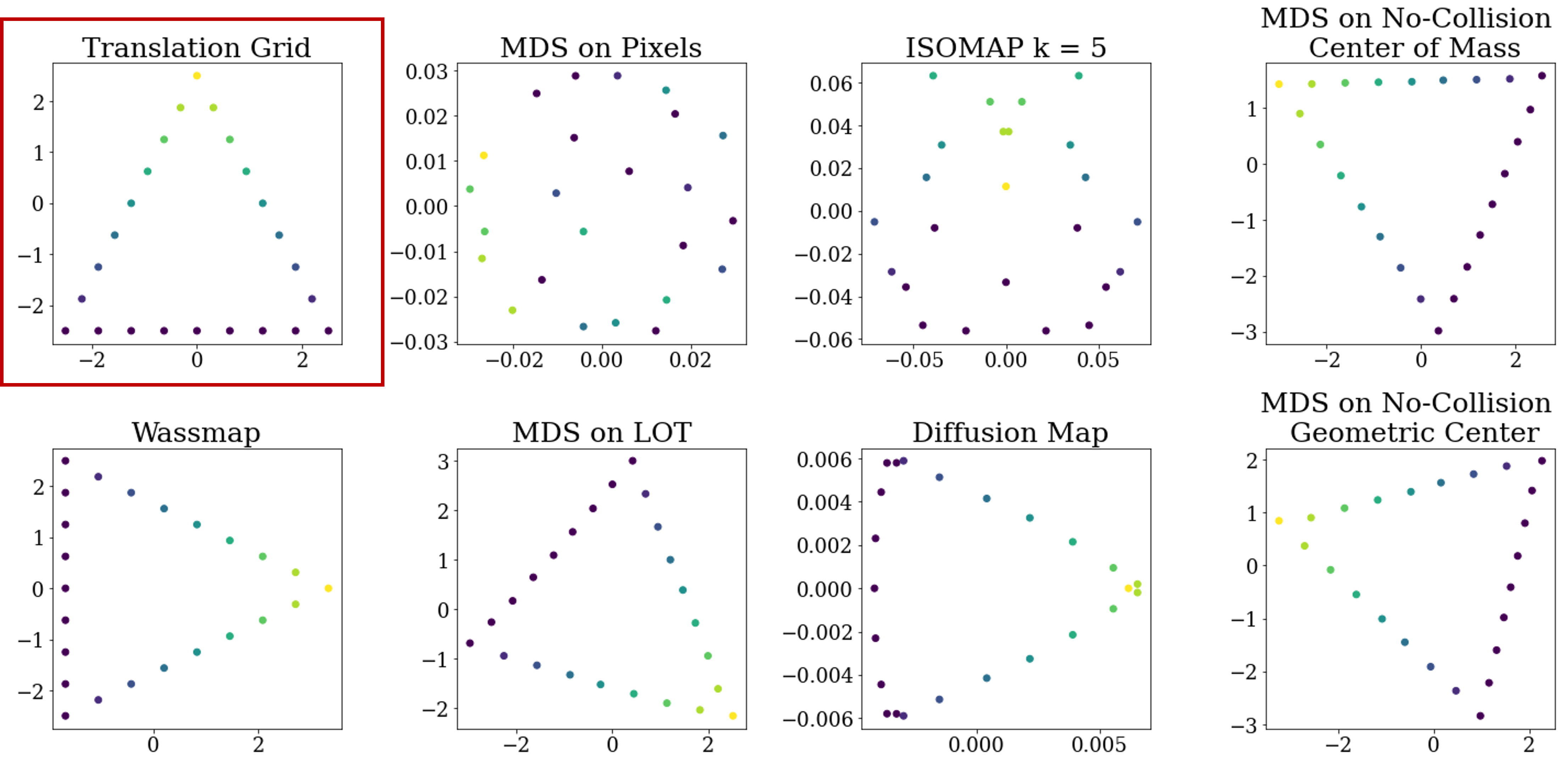}
    \caption{Translation manifold generated by the characteristic function of the unit disk on a triangular translation grid with 24 translation points. We show the original translation grid (circled in red), and the embeddings obtained by MDS, Diffusion Maps, Isomap on pixel features, Wassmap, MDS on LOT features and MDS on no-collision features.}
    \label{Transl3}
\end{figure}

Finally, In Figure \ref{Transl3} we show the translation grid obtained by MDS embedding in a $2$-dimensional space using the distance matrices provided by the different methods. The results below are in line with our theoretical results in Section \ref{TransThm}: no-collision maps are able to recover translation manifolds up to rigid transformation. The scaling is also recovered correctly after re-scaling as explained in Remark \ref{rescale}. Wassmap and LOT also produce an accurate embedding up to rigid transformation, while Isomap, Diffusion Maps and MDS on the original pixel features using Euclidean distances show significant skewing and overlapping points.
From this example we see that while all methods relying on Wasserstein distance or its approximation provide very accurate embeddings, no-collision map require only a fraction of the computational time.

\newpage
\subsection{Dilation Manifold}
We fix a base measure $\mu_0$ to be the indicator function of a disc of radius 1 centered at the origin. For a given dilation set $\Theta = \{\theta=(\theta_1,\theta_2)\} \subset \mathbb{R}^2$ the corresponding dilation manifold is given by:
$$\mathcal{M}^{\text{dil}}(\mu_0, \Theta) := \{ (\theta \odot x)\sharp\mu_0;\; \theta\in \Theta\}$$
In this example we consider $\Theta = [0.5,2]\times [0.5,4]$ and sample it on a uniform $6 \times 6$ grid.

\begin{figure}[H]
    \centering
    \includegraphics[width = \linewidth]{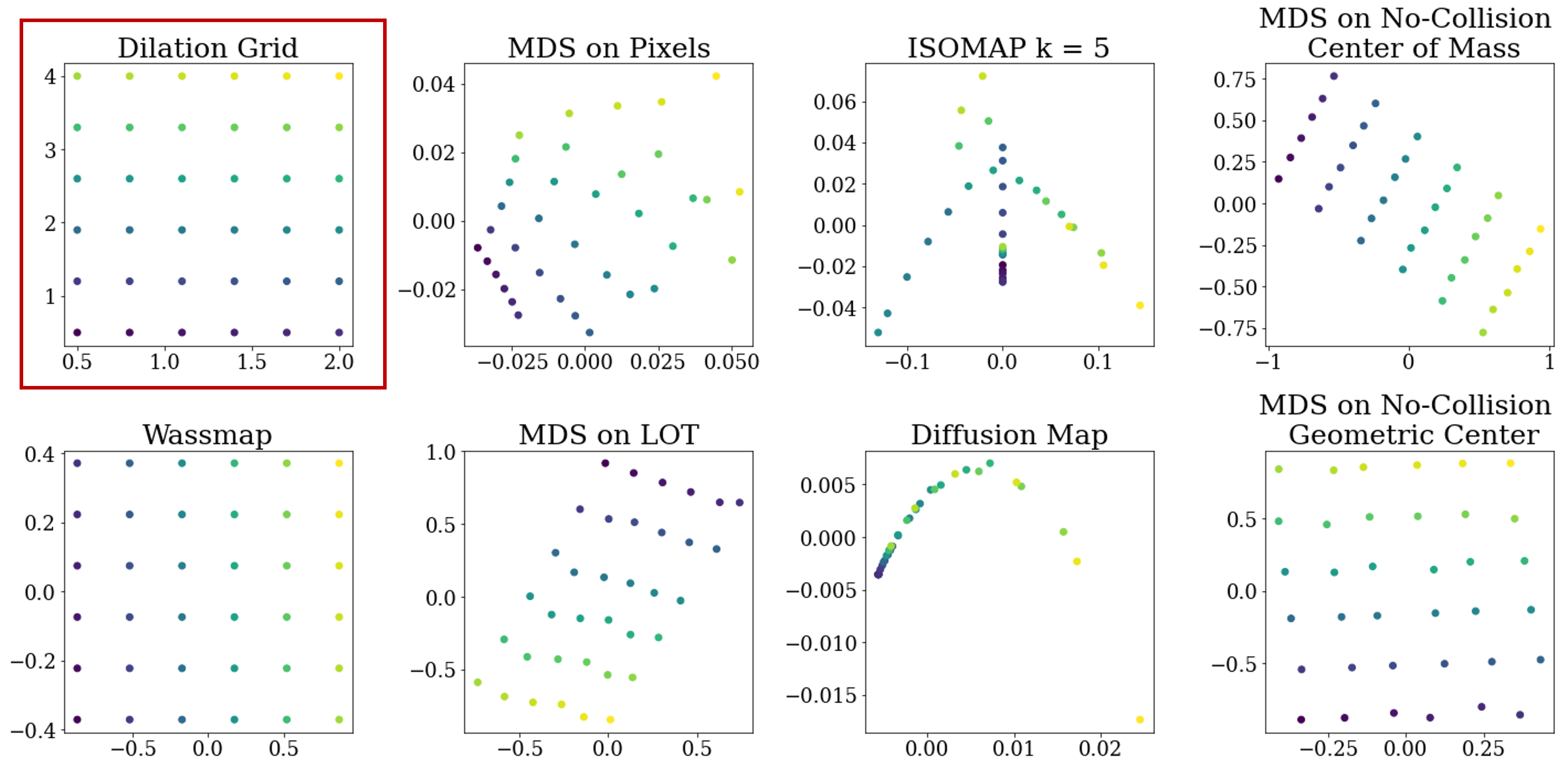}
    \caption{Dilation manifold generated by the characteristic function of the unit disk with parameter set $\Theta$ sampled on a $6\times 6$ grid. We show the original dilation grid (circled in red), and the embeddings obtained by MDS, Diffusion Maps, Isomap on pixel features, Wassmap, MDS on LOT features and MDS on no-collision features.}
    \label{Dil}
\end{figure}

As in the previous example we compute the distance matrices for the different methods and perform a 2-dimensional embedding using MDS. We use $N=3$ cuts for center of mass and $N=6$ cuts for geometric center and for LOT reference we use a Guassian centered at $(0,0)$ with covariance matrix $25I$.
In Figure \ref{Dil} we show the results of this embedding for the different methods. Again, in line with out theoretical results in Section \ref{DilThm}, no-collision maps are able to reconstruct the dilation manifold correctly modulo a rigid transformation and a scaling. Wassmap also produces an accurate embedding while LOT is able to capture the correct form of the manifold, but shows some skewing at the boundary of the grid. On the contrary MDS, Isomap and Diffusion Maps on pixel features are not able to reconstruct the underlying dilation manifold. 
Similar analysis as in the translation example can be performed on the accuracy of Wasserstein distance approximation and computational time for the different methods. As for translations, no-collision maps give a better Wassertein distance approximation than LOT and are about 500 times faster than Wassmap and 70 times faster than LOT. We refer to the Supplementary Materials for more details.

\subsection{Rotation Manifold}

As proved in Section \ref{RotThm}, in general the rotation manifold 
$$\mathcal{M}^{\text{rot}}(\mu_0, \mathbb{S}^1) = \left\{(R_{t}x)\sharp \mu_0~:~t\in [0,2\pi]\right\}$$
is not isometric to a circle. Specifically, in Section \ref{RotThm} we showed that if $a,b>0$, $a\neq b$, and $\mu_0$ is the uniform probability measure over the elliptical domain
\begin{equation*}
E_0=\left\{(x_1,x_2)\in \R^2~:~\frac{x_1^2}{a^2}+\frac{x_2^2}{b^2} \leq 1\right\}.
\end{equation*}
 and $\mu_t=(R_{t}x)\sharp \mu_0$, then $\left(\{\mu_t\}_{t\in [0,2\pi]},W_2\right)$ is not isometric to a circle. On the other hand, if $a = b$ then the two spaces are isometric.
 
\begin{figure}
    \centering
    \includegraphics[width = \linewidth]{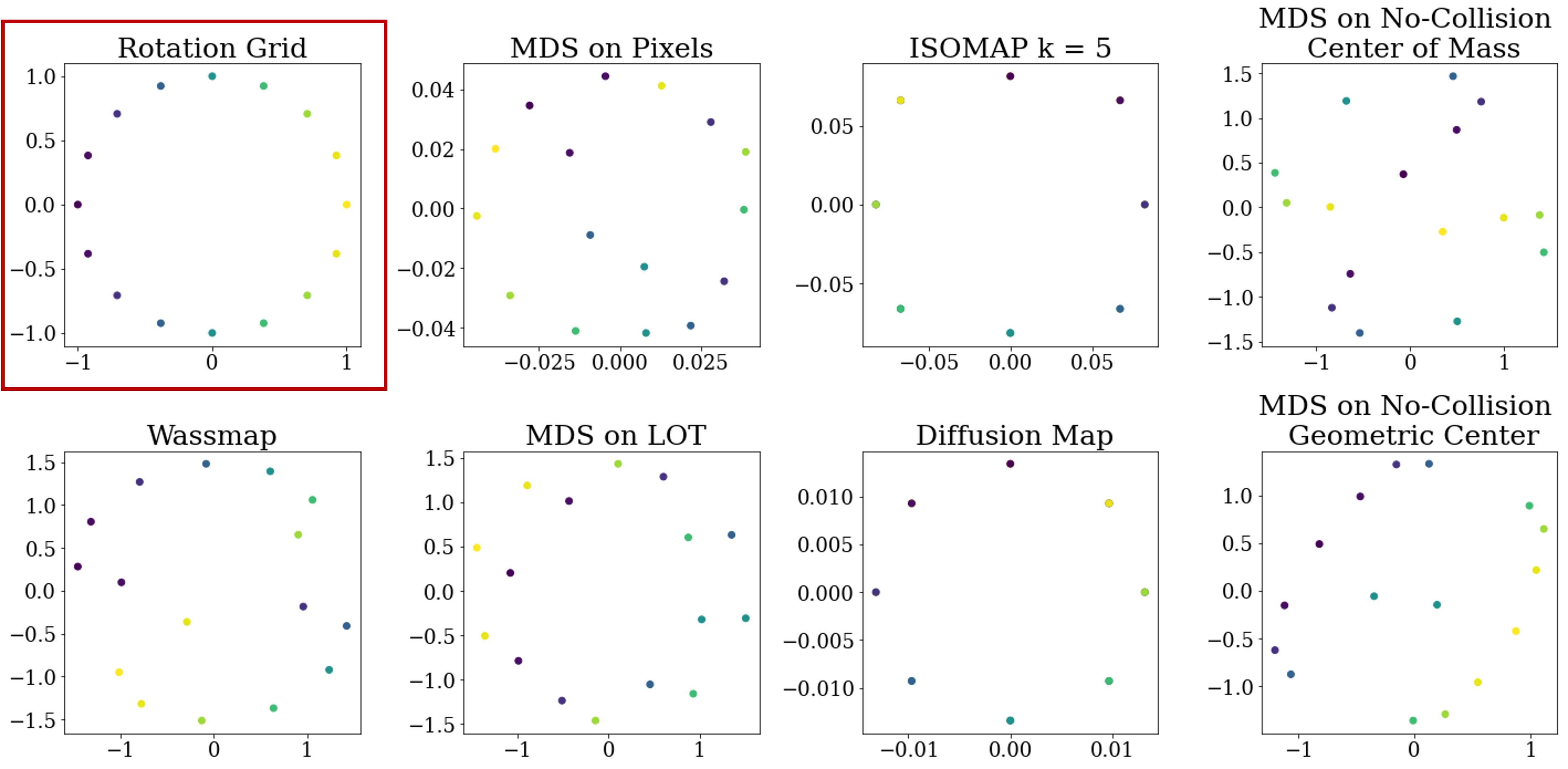}
    \caption{Rotation manifold generated an ellipse centered at $(0,1)$ with axis $a=5,\, b=2$. We show the original rotation grid (circled in red), and the embeddings obtained by MDS, Diffusion Maps, Isomap on pixel features, Wassmap, MDS on LOT features and MDS on no-collision features.}
    \label{Rota5b2}
\end{figure}

 In this example we define $\mu_0$ to be the uniform probability measure over the ellipse of center $(0,1)$ with $a =5,\; b =2$. In this case, since there is no isometry between $\left(\{\mu_t\}_{t\in [0,2\pi]},W_2\right)$ and a circle, we will see that the methods fail to reconstruct the ellipse correctly in Euclidean space. Note that this consideration only concerns the accuracy of the embedding in Euclidean space. In this case for no-collision maps we use $N=3$ cuts for center of mass and $N=8$ cuts for geometric center since these were the choices that gave the best results. In Figure \ref{Rota5b2} we show the rotation grids obtained by MDS embedding in a 2 dimensional space using the distance matrices provided by the different methods. As expected, since in this example we selected $a=5,\, b=2$ and the isometry is not satisfied, all methods based on Wasserstein distance (or an approximation of it) are unable to reconstruct the rotation grid exactly. 
 We refer to the Supplementary Materials for the distance approximation analysis and for an example in which $a = b$ and the two spaces are isometric.
 \begin{remark}\label{MDS_inst}
     (The metric) MDS, like many optimization-based algorithms, can produce slightly different results when run multiple times, especially when the embedding dimension is not sufficient for recovering an isometry. This behavior can be attributed to a few factors such as random initialization, choice of optimization algorithm, which may get stuck in local minimum, as well as numerical precision. In the supplementary material we ran 5 independent runs for this case. Four out of five times we observed that algorithms based on $W_2$ distances were not able to recover rotations, as expected from the theory. However, it can happen that a circle-like shape is a local minimum for (the metric) MDS (for one out of these five runs we obtained a circle for Wassmap embedding). We explicitly notice that this instability is only observed in the case of rotations where there is no isometry. When performing independent runs for translations and dilations the embedding is consistently correct.
 \end{remark}

\subsection{Experiments with MNIST Digits}\label{Clust_MNIST}
\begin{figure}
    \centering
    \includegraphics[width = 0.6\linewidth]{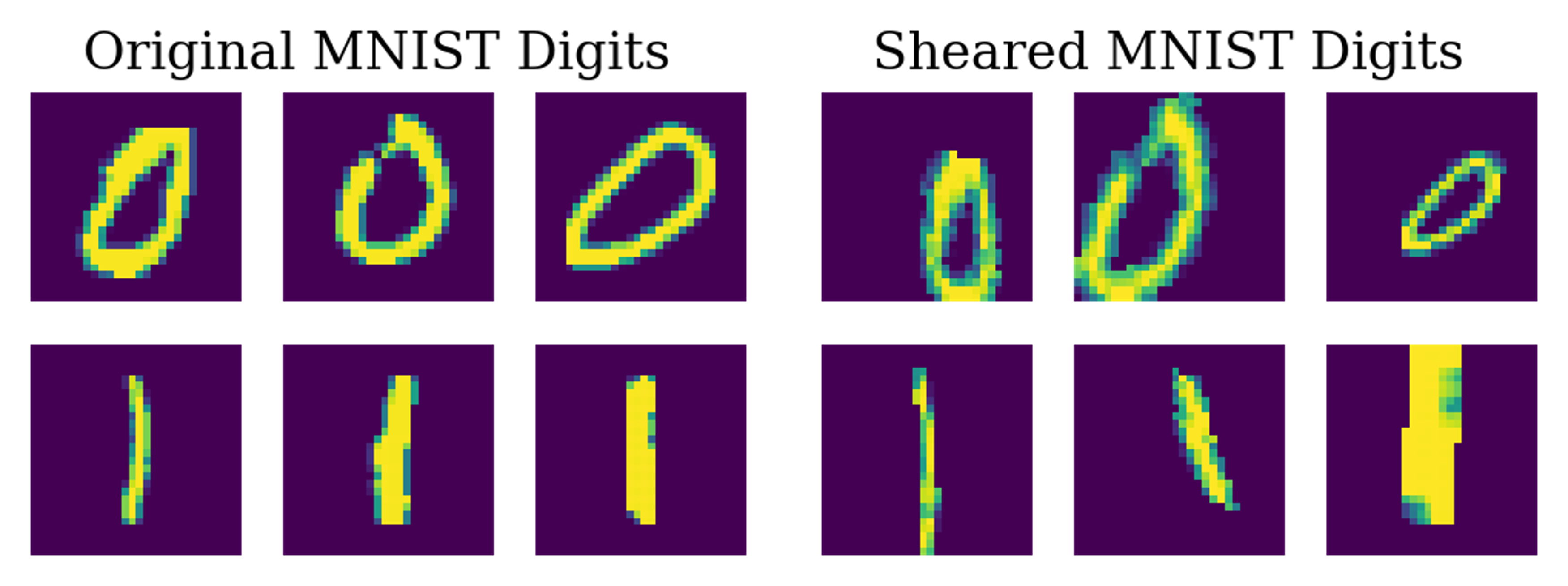}
    \caption{\textbf{Left:} Original MNIST digits. \textbf{Right:} Sheared MNIST digits.}
    \label{fig:MNIST_data}
\end{figure}

The main focus of this manuscript is on manifold learning and approximation guarantees for translations, dilations and rotations. However, since no-collision maps approximate the $W_2$ distance, it is interesting to study how this method performs for other unsupervised learning tasks on real data. In this section we propose a 2D embedding of MNIST digits \cite{deng2012mnist} using no-collision distances. As our data we use the original MNIST digits as well as sheared MNIST digits (as generated in \cite{moosmuller2023linear}). This is done to simulate a more realistic case in which the distributions are perturbed or contain noise. Example of original and sheared MNIST digits can be found in Figure \ref{fig:MNIST_data}.\\
An in-depth analysis of no-collision maps performance on clustering and classification task for more complicated datasets such as COIL, Yale, and CIFAR10 is left for a future work as we expect it to require more flexible slicing schedules.

\subsubsection{A 2D Embedding of Original MNIST Digits} 
\begin{figure}[h]
    \centering
    \includegraphics[width = 0.8\linewidth]{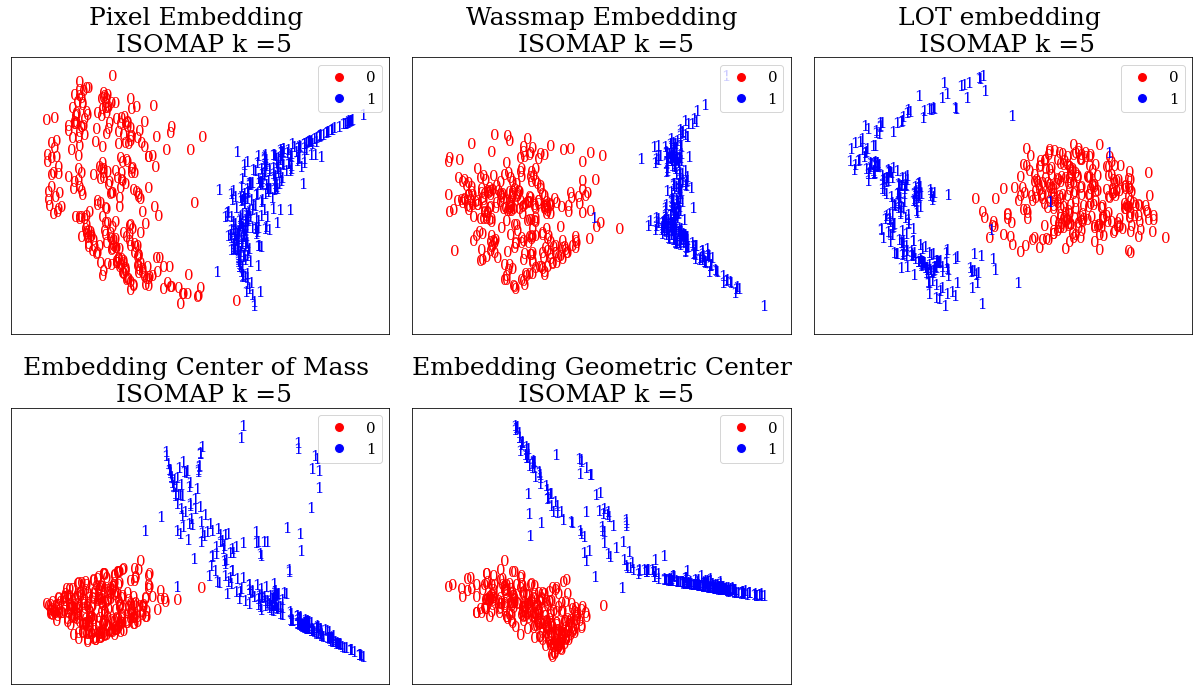}
    \caption{Embedding of original 0 and 1 digits. Digits are colored according to their label to check if digits were separated correctly.}
    \label{fig:MNIST01}
\end{figure}

Figure \ref{fig:MNIST01} shows the separation into two clusters of 600 MNIST digits ``0" and ``1". Digits are colored according to their label to check whether digits were separated correctly when using different metrics. As in previous examples, the embedding was obtained by first measuring the distances among digits (Euclidean distance between pixels, $W_2$ distance, LOT, Euclidean distance between no-collision features) then using ISOMAP with $k=5$ to produce a 2D embedding. In this case all methods provide approximately linearly separable clusters with Wassmap and no-collision providing the most linearly separable embedding with the least amount of misclassified elements. In terms of computational time, LOT and no-collision with $N=4$ cuts took respectively $9.4$ seconds and $4.4$ seconds. On the other hand, Wassmap took $902.3$ seconds.\\
We also performed a similar experiment for multi-class MNIST embedding. This can be found in the supplementary material.

\subsubsection{A 2D Embedding of Sheared MNIST Digits}\label{shearedMN}

\begin{figure}[h]
    \centering
    \includegraphics[width = 0.8\linewidth]{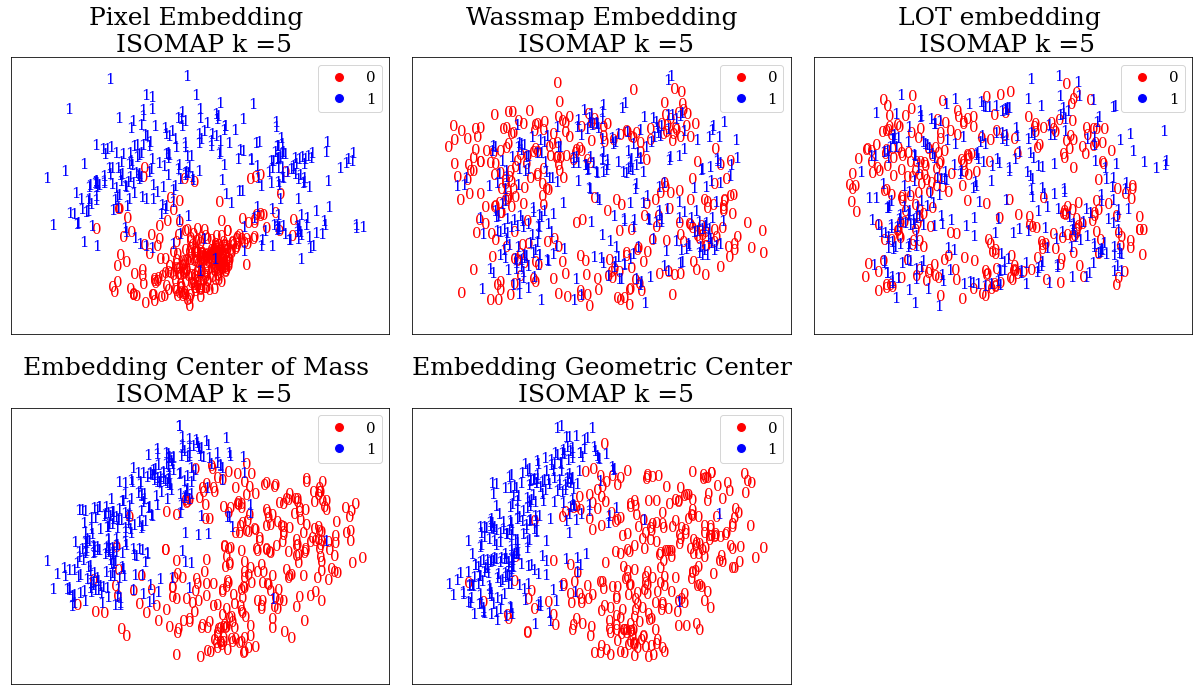}
    \caption{Embedding of sheared 0 and 1 digits. Digits are colored according to their label to check if digits were separated correctly.}
    \label{fig:sheared01}
\end{figure}

In Figure \ref{fig:sheared01} we propose a 2D embedding of 600 sheared ``0" and ``1" digits. In this case, no-collision with $N=8$ took 46.1 seconds and produced the best embedding in terms of linear separability. Wassmap and LOT did not provide a good separation in this case, while the embedding using pixel Euclidean distances provided two clusters that are not linearly separable.

\section*{Acknowledgements}
The authors also thank J. Calder, A. Cloninger, H. Kannan, C. Moosmueller, D. Slep{\v{c}}ev  and K. Hamm for valuable discussion while completing this project.

\bibliographystyle{siamplain}
\bibliography{references}
\end{document}